\documentclass{article}

\PassOptionsToPackage{numbers, compress}{natbib}

\usepackage[preprint]{neurips_2024}

\usepackage[utf8]{inputenc} 
\usepackage[T1]{fontenc}    
\usepackage{hyperref}       
\usepackage{url}            
\usepackage{booktabs}       
\usepackage{amsfonts}       
\usepackage{nicefrac}       
\usepackage{microtype}      
\usepackage{xcolor,enumitem, multirow}         

\hypersetup{
    colorlinks=true, 
    linkcolor=blue,
    filecolor=blue,      
    urlcolor=blue,
    citecolor=blue 
    }
\usepackage{amsmath}
\usepackage{amssymb}
\usepackage{mathtools}
\usepackage{amsthm}
\usepackage{bbm}
\usepackage{multicol}
\usepackage{subcaption}
\RequirePackage{algorithm}
\RequirePackage{algorithmic}

\theoremstyle{plain}
\newtheorem{theorem}{Theorem}[section]

\newtheorem{lemma}[theorem]{Lemma}

\theoremstyle{definition}
\newtheorem{definition}[theorem]{Definition}

\theoremstyle{remark}
\newtheorem{remark}[theorem]{Remark}
\allowdisplaybreaks

\newcounter{case}
\makeatletter
\newenvironment{case}[1][htb]{%
  \let\c@algorithm\c@case
  \renewcommand{\ALG@name}{Subroutine}
  \begin{algorithm}[#1]%
  }{\end{algorithm}
}
\makeatother
\title{Online Uniform Sampling:
Randomized Learning-Augmented Approximation Algorithms with Application to Digital Health}

\author{%
  Xueqing Liu \\
  Duke-NUS Medical School\\
  \texttt{xueqing\_liu@u.duke.nus.edu} \\
  \And
  Kyra Gan \\
  Cornell Tech \\
  \texttt{kyragan@cornell.edu} \\
  \And
  Esmaeil Keyvanshokooh \\
  Texas A\&M \\
  \texttt{keyvan@tamu.edu} \\
  \And
  Susan Murphy\thanks{Susan Murphy holds concurrent appointments as a Professor of Statistics and Computer Science at Harvard University and as an Amazon Scholar. This paper describes work performed at Harvard University and is not associated with Amazon.} \\
  Harvard University \\
  \texttt{samurphy@g.harvard.edu} \\
}

\begin{document}

\maketitle

\begin{abstract}
Motivated by applications in digital health, this work studies the novel
problem of \emph{online uniform sampling} (OUS), where the goal is to distribute a sampling budget uniformly across \emph{unknown} decision times. 
In the OUS problem, the algorithm is given 
a budget $b$ and a time horizon $T$, and an adversary then chooses a value $\tau^* \in [b,T]$, which is revealed to the algorithm online. 
At each decision time $i \in [\tau^*]$, the algorithm must determine a sampling probability that maximizes the budget spent throughout the horizon, respecting budget constraint $b$, while achieving as uniform a distribution as possible over $\tau^*$. 
We present the first randomized algorithm designed for this problem and subsequently extend it to incorporate learning augmentation.
We provide \emph{worst-case} approximation guarantees for both algorithms, and
illustrate the utility of the algorithms through both
synthetic experiments and a real-world case study involving the HeartSteps mobile application.
Our numerical results show strong empirical \emph{average} performance of our proposed randomized algorithms against previously proposed heuristic solutions.
\end{abstract}

\section{Introduction}
The problem of \emph{online uniform sampling} (OUS)
is motivated by applications in digital health, where administering interventions at inappropriate
times,
such as when users are not at risk,\footnote{Risk times are when the patient is susceptible to a negative event, such as smoking relapse.}
can significantly increase mental burden and hinder engagement with digital interventions \citep{li2020microrandomized, nahum2018just, wen2017compliance, mcconnell2017feasibility, mann2009boredom}. 
Existing studies \citep{heckman2015treatment,klasnja2008using,dimitrijevic1972habituation} show excessive digital interventions can heighten user fatigue, suggesting a threshold beyond which intervention effectiveness declines.
A strategy rooted in the ecological momentary assessment (EMA) literature and
proven effective in mitigating user fatigue
\emph{involves allocating a fixed and limited budget for treatments delivered to the patient and delivering them with a uniform distribution across all risk times} (e.g., \citealt{liao2018a, dennis2015pilot, rathbun2013survival, scott2017pilot, scott2017using, shiffman2008ecological, stone2007science}). However, this strategy is challenging because the true number of risk times is unknown, inspiring the OUS problem.

\textbf{Contributions}
Our contributions in this paper are two-fold. \textbf{First}, we formulate the common OUS problem in digital health as an online optimization problem and provide randomized algorithms that perform well in practice with \emph{competitive ratio} guarantees.
The competitive ratio measures
the performance of an online algorithm against
an offline clairvoyant benchmark, assuming
the unknown parameter is revealed to the clairvoyant in advance.
These guarantees are inherently conservative:
1) no online algorithm can achieve the same performance as the clairvoyant in practice (i.e., a competitive ratio of 1  is unattainable in OUS), and 2) they hold across \emph{all} problem instances or sample paths (i.e., they are worst-case guarantees). 
Consequently, online approximation algorithms may exhibit conservative behavior. To address this, we numerically illustrate the practicality of our algorithm, demonstrating that they outperform naive benchmarks on average.
\textbf{Second}, we extend our algorithm to the practical setting where a confidence interval \emph{containing} the true risk time is provided, potentially through a valid
statistical inference procedure.
We conduct the competitive ratio analysis for our proposed learning-augmented approximation algorithm, demonstrating its {\em consistency} in the strong sense---optimal performance is achieved when the confidence interval width is zero---and {\em robustness}---the learning-augmented algorithm performs no worse than the non-learning augmented counterpart. Our findings indicate that, in almost all tested scenarios, the randomized learning-augmented algorithm outperforms its non-learning augmented counterpart.


\textbf{Outline}
In Section  \ref{problem},
we 
formalize the OUS problem.
We introduce our randomized algorithm without learning augmentation
in Section \ref{alg}. 
This algorithm is segmented into three distinct cases based on the horizon length to budget ratio, with a competitive ratio established for each.
In Section \ref{alg-learn},
we develop a learning-augmented algorithm that integrates a prediction interval and provide theoretical justification for its effectiveness. The efficacy of these algorithms is 
first assessed through synthetic experiments, followed by their application to real-world data
in Section \ref{exp}.

\vspace{-8pt}
\subsection{Related Work}\label{subsec:related_work}

\vspace{-5pt}
\textbf{Online Uniform Sampling$\;$}
Existing methodologies, primarily sourced from the EMA literature, 
focus on delivering interventions through the form of mobile self-report requests over a fixed horizon, constrained by budget and uniformity considerations to minimize user burden and accurately reflect user conditions across diverse contexts \citep{dennis2015pilot,rathbun2013survival,scott2017pilot,scott2017using}. 
In this work, we permit intervention only when users are \emph{at risk}, leading to an unknown horizon length, which creates a significant challenge in balancing the maximum budget spent with uniformity considerations. 
To address this issue, \citet{liao2018a} developed a heuristic algorithm, but its performance depends
heavily on the accuracy of the predicted number of risk times.
When the prediction is inaccurate, the algorithm lacks theoretical guarantees, highlighting the need for a more robust algorithm design.

\textbf{Multi-option Ski-rental Problem}
Our work  closely relates
to the \emph{multi-option ski-rental} (MOSR) problem \citep{zhang2011ski,shin2023improved}, where the number of snowy days is unknown. 
Customers have
multiple ski rental options, differing in cost and duration. The goal
is to minimize costs while ensuring ski availability on snowy days. \citet{shin2023improved} introduced a
randomized 
algorithm for MOSR,
with a tight $e$-competitive ratio. 
A random variable $B$ is introduced as a proxy for 
the unknown true horizon $T$. 
$B$ is initialized
to $\alpha$, following a density function $1/\alpha$ within $[1,e)$.  The algorithm iteratively solves an optimization problem to identify an optimal set of rental options within budget $B$, maximizing day coverage. Customers sequentially utilize the options until depletion, at which point $B$ is increased by a factor of $e$, and the process is repeated. 

Our work builds upon \citet{shin2023improved}, leveraging the same randomized algorithmic idea. However, our problem setting is \emph{significantly different} from that of MOSR. In particular, instead of having discrete ski-rental options, at each decision time, the algorithm needs to decide on the sampling probability, which is continuous in nature. Further, in our problem, the sum of the sampling probability cannot exceed a predefined budget, while such constraints do not exist in MOSR. Our problem additionally has a uniformity consideration.



\textbf{Learning-Augmented Online Algorithms$\;$}
Many online algorithms 
incorporate black-box point predictions on the unknown parameters
to improve their worst-case 
guarantees
\citep{purohit2018improving, bamas2020primal, wei2020optimal, jin2022online}.
The confidence of these point estimates is often represented by a single parameter, with a higher value indicating more accurate predictions.
When the confidence is low,
most work
do not guarantee that the learning-augmented algorithm will perform no worse 
than the non-learning counterpart \citep{bamas2020primal}.
Often, in practice, prediction confidence intervals, rather than point estimates, can be generated by a valid statistical inference method, where a wide confidence interval corresponds to less informative predictions  \citep{shafer2008tutorial}. 
\citet{im2021online} 
consider the setting 
where the prediction provides a range of values for key parameters in the online knapsack problem.
However, their
deterministic
solution
cannot be extended into our setting since 
the number of risk times is stochastic in OUS.
We present the first integration of prediction intervals to
OUS,
enabling our proposed algorithms to surpass the performance of their non-learning counterparts even with a wide
prediction interval.




\vspace{-5pt}
\section{Problem Framework}
\vspace{-3pt}
\label{problem}
In the context of digital interventions, we define the OUS problem as presented by \citet{liao2018a}.
Let $T$ denote the maximum number of decision points within a decision period (e.g., within a day).
At any given time $t\in[1, T]$ in each decision period, patients encounter binary risk levels\footnote{When multiple risk levels are present, the problem naturally decomposes into independent subproblems for each risk level, see more details in Section \ref{app:mult} of the Supplementary Material.} (determined by data  from wearable devices), indicating whether the patient is likely to experience an
adverse event, such as relapse to smoking. 
The distributions of risk level are allowed to change \emph{arbitrarily} across decision periods since  the treatment may reduce subsequent risk.

Let $\tau^*$ be the \emph{unknown} \emph{true} number of risk times that a patient experiences in a decision period.
Note that 
$\tau^*$ is stochastic and is revealed \emph{only} at the end of the horizon $T$, corresponding to the last decision time in the decision period. 
We define $p_{i}\in (0,1)$ to be the treatment probability at  time $i\in[\tau^*]$. 
We preclude the possibility that $p_{i}=0$ or $p_{i}=1$ to facilitate after-study inference~\citep{boruvka2018assessing, zhou2023offline, kallus2022stateful}.

The algorithm is provided with a \emph{soft} budget of $b$, representing the total \emph{expected} number of interventions allowed to be delivered within each decision period. We assume $\tau^*> b$ 
as evidenced in practice~\citep{liao2018a}.
At each decision time $i$, the algorithm
decides  
the intervention probability $p_{i}$.
The objectives of the OUS problem \citep{liao2018a, dennis2015pilot, rathbun2013survival, scott2017pilot, scott2017using, shiffman2008ecological, stone2007science} are to
1) assign the intervention probabilities $\{p_{i}\}_{i\in[\tau^*]}$  as uniform as possible 
across risk times, and 2) maximize the sum of intervention probabilities across risk times while adhering to the budget constraint $b$. 
 
Abstractly, in OUS,
the algorithm is given a budget $b$ and a time horizon $T$, and an adversary then chooses a value $\tau^* \in [b,T]$, which is revealed to the algorithm online. At each decision time $i \in [\tau^*]$, the algorithm must determine a sampling probability that maximizes the budget spent throughout the horizon, respecting the budget constraint $b$, while achieving as uniform a distribution as possible over $\tau^*$.

\emph{Without additional information on $\tau^*$}, the two objectives compete with each other. 
A naive solution 
to fulfill
the first objective is to set $p_{i} = b/T, i \in [\tau^*]$, which, however,  fails to
maximize the sum of intervention probabilities. Conversely, if we set $p_{i}$ to be a large value
and keep it constant,
there is a risk of depleting the budget before the end of the horizon, thus failing to achieve the uniformity objective.
Therefore, the optimality of the two objectives cannot be simultaneously achieved 
without
additional information on $\tau^*$. \citet{liao2018a} provided a heuristic algorithm for OUS
given 
a point estimate of $\tau^*$. The algorithm's performance is significantly influenced by the accuracy of this forecast.
In this work, we introduce randomized algorithms for OUS with robust worst-case guarantees, considering settings both with and without learning augmentation.


\subsection{OUS as An Online Optimization Problem}
\label{ous}
In this section, we formulate OUS formally as an online optimization problem, where the objective function provides a uniform way of comparing the performance of different approximation algorithms, and the constraint defines the set of feasible solutions. 



Formally, we aim to find a sequence of 
$\{p_{i}\}_{i\in[\tau^*]}$ that achieves the following two objectives: 
\vspace{-10pt}
\begin{enumerate}[leftmargin=*, itemsep=0pt, parsep=0pt]
    \item Maximizes the sum of treatment probabilities across risk times, subject to the ``soft'' budget $b$;
\item Penalizes changes in treatment probabilities within each risk level.
\end{enumerate}
\vspace{-10pt}
The OUS can be expressed using the
following optimization problem:
\begin{align}\label{raw_opt}
\Bigg\{
\max &\sum_{i}^{\tau^*} p_{i} - \frac{1}{\tau^*}\ln\left(\frac{\max_{i\in[\tau^*]}p_{i}}{\min_{i\in[\tau^*]}p_{i}}\right): \nonumber \\
    &\mathbb{E}\Bigg[\sum_{i=1}^{\tau^*}
    p_{i}\Bigg]  \leq b, 
    p_{i}
    \in (0,1),  \forall i\in[\tau^*].\Bigg\}
    \vspace{-10pt}
\end{align}
where the expectation, $\mathbb{E}$, in the budget constraint is taken over the randomness in the algorithm. This budget constraint is ``soft'' in the sense that if we have multiple decision periods (which is the case in digital health), we should satisfy the budget constraint in expectation.


\begin{remark}
Notably, the purpose of formulating the optimization problem is not to solve it optimally, but rather to provide a feasible solution without knowledge of the unknown $\tau^*$.
Rather than setting uniformity as a constraint, we incorporate it into the design of our approximation algorithms.
By including uniformity as a penalty term in the objective function, represented by:
%
%
\vspace{-5pt}
\begin{equation}\label{eq:penalty}
\frac{1}{\tau^*} \ln\left(\frac{\max_{i\in[\tau^*]}p_{i}}{\min_{i\in[\tau^*]}p_{i}}\right),
\vspace{-5pt}
\end{equation}
we can directly compare the overall performance of different online approximation algorithms, including how well they achieve uniformity, by comparing their objective function values.

The choice of the penalty term \eqref{eq:penalty} is inspired by the entropy change concept from thermodynamics~\citep{smith1950introduction}. This choice is not unique but it has several nice properties: 
a) it equals to 0 if and only if $\{p_{i}\}_{i\in[\tau^*]}$ are identical,  b) it increases with the maximum difference in $\{p_{i}\}_{i\in[\tau^*]}$, and c) it tends towards infinity as the value of $p_{i}$ approaches to zero, penalizing scenarios where the expected budget is depleted before the horizon ends. We note that one can replace the term $1/\tau^*$ in the penalty by a tuning parameter $\sigma$, which controls the strength of the penalty, as discussed in Remarks \ref{remark: th3.2} and \ref{remark: th4.2}.\footnote{Since the current design of our algorithms does not explicitly account for the form of the penalty term, the penalty \eqref{eq:penalty} could also be replaced by any other suitable functions, with performance re-evaluated under the modified objective function. }
Finally, we highlight that KL divergence cannot be used here to impose uniformity (see detailed discussion in Appendix~\ref{app:penalty}).
\end{remark}


\subsection{Offline Clairvoyant and Competitive Ratio}
In the \emph{offline clairvoyant} benchmark, the clairvoyant possesses knowledge of $\tau^*$.
When provided with this value, the optimal solution of Problem~\eqref{raw_opt} is to set $p_{i} = b/\tau^*$.
Then, the optimal value of the  objective function 
in Problem~\eqref{raw_opt} is $\text{OPT}(\tau^*)=b$.
Importantly, 
in practice, no online algorithm can attain $\text{OPT}(\tau^*)$.
Let $\text{SOL}$ be the objective value of Problem~\eqref{raw_opt} achieved by a \emph{randomized online} algorithm, we say that

\begin{definition}[$\gamma$-competitive]\label{def:competitive}
An algorithm is $\gamma$-\textit{competitive} if $\mathbb{E}[\text{SOL}] \geq \gamma \cdot \text{OPT}(\tau^*)$. 
\end{definition}

\begin{remark}\label{remark:expectation}
First, we highlight that the expectation in Definition~\ref{def:competitive} is \emph{only} taken over the randomness of the algorithm. 
Second, we note that if provided, the competitive ratio would hold for every feasible $\tau^*\in[b,T]$ in expectation.  This implies that competitive ratio is a worst-case guarantee: in an OUS instance
so long as the budget $b$ and maximum horizon length $T$ are fixed across decision periods, we can expect to meet the budget and achieve the above competitive ratio regardless of the specific realization of $\tau^*$ in each decision period.
\end{remark}
The key difficulty in solving Problem~\eqref{raw_opt} in the online setting arises due to the unknown nature of $\tau^*$. 
Inspired by the randomized algorithm proposed by \citet{shin2023improved}, 
%
we introduce the first approximation algorithm for the OUS problem. 

\vspace{-5pt}
\subsection{With Learning Augmentation}
In the \emph{learning-augmented} setting,
we are additionally given a prediction confidence interval $[L,U]$, generated by a valid statistical procedure, that contains
the unknown \emph{true} $\tau^*$ with high probability. A wider confidence interval indicates lower prediction quality.
For simplicity, 
we assume $\tau^*$ is within the interval, 
though our results generalize to cases where it is contained with high probability.
To measure 
the performance of the learning-augmented algorithm
 in the presence of a prediction confidence interval, we extend the standard definition of consistency-robustness analysis from previous literature \citep{lykouris2021competitive,purohit2018improving,bamas2020primal, shin2023improved}.
Specifically, an algorithm is 
said to be
$\lambda$-\textit{consistent} 
if it achieves  $\mathbb{E}[\text{SOL}] \geq \lambda \cdot \text{OPT}(\tau^*)$ when the prediction is perfect, i.e., 
when $L=U$, indicating a zero-length interval.\footnote{Similar to Definition~\ref{def:competitive}, the expectation is taken over the randomness in the algorithm.} This corresponds to the previous definition where the prediction is accurate \citep{shin2023improved}.
Conversely, we say an algorithm is $\rho$-\textit{robust}
if it satisfies $\mathbb{E}[\text{SOL}] \geq \rho\cdot \text{OPT}(\tau^*)$ regardless of the width of the prediction interval $[L,U]$ (with a wider confidence interval indicating a less informative prediction).  This corresponds to the previous definition where the prediction can be arbitrarily inaccurate. 
In Section~\ref{alg-learn}, we establish that our proposed learning-augmented algorithm is $1$-consistent,
achieving the optimal solution when the interval width is zero, and the competitive ratio of our learning-augmented algorithm is close to that of the non-learning augmented counterpart. To the best of our knowledge, this is the first work that provide a 1-consistency guarantee on learning-augmented algorithms, after careful engineering of the algorithms.

\vspace{-5pt}
\section{Randomized Algorithm}
\label{alg}
In this section, we introduce our randomized algorithm, Algorithm~\ref{alg:rand}, designed for the OUS problem {\em without} learning augmentation. 
This algorithm is inspired by the randomized algorithm proposed by \citet{shin2023improved} for the MOSR problem. Due to the significant differences in problem setup outlined
in Section~\ref{subsec:related_work}, the design of our algorithm requires 1) imposing a discrete structure on the sampling probabilities to account for uniformity considerations, making the analysis of the algorithm more tractable, and 2) explicitly addressing the finite horizon length and budget constraint, ensuring that the randomized algorithm does not exceed the budget in expectation.

\begin{algorithm}[ht!]
\caption{Randomized Online Algorithm }
\label{alg:rand}
\begin{algorithmic}[1]
\STATE \textbf{Input:} $T$, $b$ 
\STATE \textbf{Initialize:}  $j=1$, we sample $\alpha \in [b,be]$ from a distribution with p.d.f. $f(\alpha) = 1/\alpha$, 
and initialize $\tilde{\tau} = \alpha$
\FOR{$i=1,...,\tau^*$}
\STATE  We calculate: 
\[\mathrm{Int}(\tilde{\tau}) = \left\{
    \begin{array}{ll}
      \lfloor\tilde{\tau} \rfloor   &  w.p. \quad \lceil\tilde{\tau} \rceil - \tilde{\tau}  \\
       \lceil\tilde{\tau} \rceil   &  w.p. \quad \tilde{\tau} - \lfloor\tilde{\tau}\rfloor
    \end{array}
    \right.\]
\IF{$T \leq    b e$}
\STATE{Update $\tilde{\tau}$ and set $p_{i}$ using \textbf{Subroutine}~\ref{alg:rand_case1}}

\ELSIF{$   b e < T \leq    b e^2$}
\STATE Update $\tilde{\tau}$ and set $p_{i}$ using \textbf{Subroutine}~\ref{alg:rand_case2}
\ELSE 
\STATE{Update $\tilde{\tau}, b$ and set $p_{i}$ using \textbf{Subroutine}~\ref{alg:rand_case3}}
\ENDIF
\STATE Output treatment probability $p_{ i}$
\ENDFOR
\end{algorithmic}
\end{algorithm}


The proposed algorithm, Algorithm~\ref{alg:rand},  provides a feasible solution to Problem~\eqref{raw_opt}. At its core, our algorithm assigns the sampling probabilities in a monotonically non-increasing fashion over time. 
To accommodate varying practical scenarios where the budget-to-horizon ratio differs across applications, 
we designed specialized approximation algorithms for three possible scenarios:
1) $T\leq b e$ (\textbf{Subroutine}~\ref{alg:rand_case1}), 2) $b e<T \leq b e^2$ (\textbf{Subroutine}~\ref{alg:rand_case2}), and 3) $T>b e^2$ (\textbf{Subroutine}~\ref{alg:rand_case3}).

We maintain a running  ``guess'' of $\tau^*$, denoted by $\tilde\tau$. We initialize $\tilde\tau$ to be $\alpha$, where $\alpha \sim[b, b\cdot e]$ with density $1/\alpha$, and $e$ represents the Euler's number.
If the current number of risk times
$i$ is within our running guess $\tilde\tau$, then we do not change the current sampling assignment probability. 
Otherwise, we update 
$\tilde \tau$ as $\tilde \tau = \tilde \tau e $
and update the sampling probability according to Algorithm~\ref{alg:rand}, depending on the length of the horizon $T$ relative to $b$. The random draw $\tilde \tau$ controls not only the value of the sampling probability but also the duration of each stage. Once the algorithm reaches $\tilde \tau$, it transitions to the next stage, resulting in a stage-wise constant probability sequence. 

We first show the feasibility of our proposed solution, i.e., the sampling probabilities outputted from Algorithm~\ref{alg:rand} satisfies the budget constraint in Problem~\eqref{raw_opt}:
\begin{lemma}\label{lemma:feasibility_rand}
Let $p_i^{A1}$ be the probability returned by Algorithm  \ref{alg:rand} at risk time $i \in [\tau^*]$.
    This solution always satisfies the budget constraint in expectation, $\mathbb{E}\left[\sum_{i=1}^{\tau^*}
    p_{i}^{A1}\right]  \leq b$, where the expectation is taken over the randomness of the algorithm.
\end{lemma}

Next, 
by leveraging the monotonically non-increasing nature of the sampling probabilities, 
the objective in Problem~\eqref{raw_opt} simplifies to 
\begin{equation}\label{eq:alg_obj}
    \max \sum_{i=1}^{\tau^*} p_{i} -  \frac{1}{\tau^*}\ln\left(\frac{p_{1}}{p_{\tau^*}}\right).
\end{equation}
Using Equation~\eqref{eq:alg_obj}, 
we compute the competitive ratio of Algorithm~\ref{alg:rand}:
\begin{theorem}\label{theorem:cr_rand}
Algorithm~\ref{alg:rand} is $\mathcal{X}(T)$-competitive, where $\mathcal{X}$ is defined as follows:
    \begin{align*}
        \mathcal{X}(T) := \left\{ \begin{array}{ll}
          \frac{1}{e}\left(\ln (e-1)+\frac{1}{e-1}\right)   & \text{if} \quad  T \leq b e, \\
           \frac{1}{e}  & \text{if} \quad b e < T \leq b e^2,\\
          \frac{1}{e} - \frac{1}{e^2} & \text{if} \quad T > b e^2.
        \end{array} \right.
    \end{align*}
\end{theorem}

\begin{case}[t]
\caption{($i$, $b$, $\tilde{\tau}$, $T$, $\mathrm{Int}(\tilde \tau)$)}
\label{alg:rand_case1}
\begin{algorithmic}[1]
\IF{$i > \mathrm{Int}(\tilde{\tau})$}
\STATE $\tilde{\tau} =  \tilde{\tau}e$
\ENDIF
\STATE $p_{i} = \frac{b}{\min(T, \tilde{\tau} (e-1))}$
\end{algorithmic}
\end{case}
\begin{case}[t]
\caption{($i$, $b$, $\tilde{\tau}$,  $\mathrm{Int}(\tilde \tau)$)}
\label{alg:rand_case2}
\begin{algorithmic}[1]
\IF{$i > \mathrm{Int}(\tilde{\tau})$}
\STATE $j = j+1$, $\tilde{\tau} =  \tilde{\tau}e$
\ENDIF
\IF{$j\geq 3$}
\STATE $p_{i} = \frac{b}{\tilde{\tau} e}$
\ELSE
\STATE $p_{i} = \frac{b}{\tilde{\tau} (e-1)}$
\ENDIF
\end{algorithmic}
\end{case}
\begin{case}[t]
\caption{($i$, $b$, $\tilde{\tau}$, $\mathrm{Int}(\tilde \tau)$)}
\label{alg:rand_case3}
\begin{algorithmic}[1]
\IF{$i > \mathrm{Int}(\tilde{\tau})$}
\STATE $j = j+1$, $\tilde{\tau} =  \tilde{\tau}e$
\IF{$j\geq 3$}
\STATE $b = b(1-\frac{1}{e})$
\ENDIF
\ENDIF
\STATE $p_{i} = \frac{b}{\tilde{\tau} e}$
\end{algorithmic}
\end{case}

The above competitive ratio is conservative by design: it was derived by taking the worst-case 
over \emph{unknown} $\tau^*$  and horizon length $T$ within each case.
The proof of Theorem \ref{theorem:cr_rand} in Section \ref{app:cr_rand} of the Supplementary Material outlines the competitive ratio as a function of $\tau^*$ and $T$. 
In addition, in Section~\ref{exp}
we explore the impact of varying $\tau^*$ while keeping the horizon length fixed, providing a numerical illustration of how the expected competitive ratio changes.
We note that the expected competitive ratio, averaged over the unknown $\tau^*$, is much better than our theoretical competitive ratio illustrated above. {Based on our theoretical competitive ratio in Theorem \ref{theorem:cr_rand}, we recommend choosing the horizon length $T$ relative to the budget $b$ to be below $b e^2$, which aligns with our empirical results in Section \ref{exp} (see Remark \ref{remark:design_choice_T_no_augmentation} for details).}  

\begin{remark}
\label{remark: th3.2}
As stated in Section \ref{ous}, the term $\frac{1}{\tau^*}$ in the penalty can be replaced by a tunable strength parameter $\sigma$. In Section \ref{app:cr_rand}, we show that for $T \leq be^2$, the above results hold across a wide range of $\sigma$ values, specifically $\sigma \leq \frac{b}{2}$. However, when $T > be^2$, $\sigma$ should be on the order of $\frac{1}{\tau^*}$, ensuring that the penalty term scales similarly to the budget term in the objective. 
\end{remark}

\section{Learning-Augmented Algorithm}
\label{alg-learn}
\begin{algorithm}[t!]
\caption{Randomized Online Algorithm With Prediction Confidence Intervals}
\label{alg:pred}
\begin{algorithmic}[1]
\STATE \textbf{Input:} $T$, $b$, $[L, U]$
\STATE \textbf{Initialize:} $j=1$, sample $\alpha \in [b, be]$ from a distribution with p.d.f. $f(\alpha) = 1/\alpha$, and initialize $\tilde{\tau} = \alpha$
\FOR{$i=1,...,\tau^*$}
\STATE  We calculate:

\[\mathrm{Int}(\tilde{\tau}) = \left\{
    \begin{array}{ll}
      \lfloor\tilde{\tau} \rfloor   &  w.p. \quad \lceil\tilde{\tau} \rceil - \tilde{\tau}  \\
       \lceil\tilde{\tau} \rceil  &  w.p. \quad \tilde{\tau} - \lfloor\tilde{\tau} \rfloor
    \end{array}
    \right.\]
\IF{$U \leq   b e$}
\STATE Update $\tilde{\tau}$ and set $p_{i}$ using \textbf{Subroutine}~\ref{alg:aug_case1}

\ELSIF{$b e < U \leq   b e^2$}
\IF{$U - L \leq b (e-1)$}
\STATE Update $\tilde{\tau}$ and set $p_{i}$ with \textbf{Subroutine}~\ref{alg:aug_case1}
\ELSE
\STATE Update $\tilde{\tau}$ and set $p_{i}$ with \textbf{Subroutine}~\ref{alg:rand_case2}
\ENDIF
\ELSE 
\IF{$U - L \leq b (e+1)$}
\STATE Update $\tilde{\tau}$ and set $p_{i}$ with \textbf{Subroutine}~\ref{alg:aug_case2}
\ELSE
\STATE Update $\tilde{\tau}, b$ and set $p_{i}$ with \textbf{Subroutine}~\ref{alg:aug_case3}
\ENDIF

\ENDIF
\STATE Output sampling probability $p_{i}$
\ENDFOR
\end{algorithmic}
\end{algorithm}

In this section,
we propose a new approximation algorithm, Algorithm~\ref{alg:pred}, under the learning-augmented setting,
where we are provided with prediction confidence intervals $[L,U]$ for the unknown $\tau^*$. 
Algorithm \ref{alg:pred} builds upon the non-learning augmented counterpart, 
Algorithm \ref{alg:rand}, utilizing the given confidence interval for optimization. 
Similar to Algorithm~\ref{alg:rand}, we initialize $\alpha\sim [b, b e]$ with density $1/\alpha$, and the current ``guess'' of $\tau^*$ is reflected by $\tilde\tau + L$.

In Algorithm~\ref{alg:pred}, the three scenarios differ from those in Algorithm~\ref{alg:rand}. Here, the distinction is based on the relationship between the upper bound of the interval, $U$, and the budget $b$.
The three scenarios are 
1) $U \leq b e$ (\textbf{Subroutine} \ref{alg:aug_case1}), 2) $be < U \leq b e^2$, further divided into 2a) $ U-L \leq b(e-1)$ (\textbf{Subroutine} \ref{alg:aug_case1}), and 2b) $U - L > b(e-1)$ (\textbf{Subroutine} \ref{alg:rand_case2}), and 3) $U > b e^2$,  further divided into 3a) $U-L \leq b(e+1)$ (\textbf{Subroutine} \ref{alg:aug_case2}), and 3b) $U - L > b(e+1)$ (\textbf{Subroutine} \ref{alg:aug_case3}).



Similarly, we first illustrate that Algorithm~\ref{alg:pred} produces a feasible solution to Problem~\eqref{raw_opt} and then provide a theoretical guarantee on its performance.

\begin{lemma}\label{lemma:feasibility_learn}
Let $p_i^{A2}$ be the probability returned by Algorithm  \ref{alg:pred} at risk time $i \in [\tau^*]$.
    This solution always satisfies the budget constraint in expectation, $\mathbb{E}\left[\sum_{i=1}^{\tau^*}
    p_{i}^{A2}\right]  \leq b,$ where the expectation is taken over the randomness of the algorithm.
\end{lemma}

\begin{theorem}\label{theorem:cr_learn}
    This algorithm is $1$-consistent and $\mathcal{X}(U)$-robust, where $\mathcal{X}(U)$ is defined as follows:
    \begin{align*}
        \mathcal{X}(U) := \left\{ \begin{array}{ll}
          \ln2 + \frac{e-1}{e}\ln\frac{e-1}{e}   & \text{if} \quad  U \leq b e, \\
           \frac{1}{e}  & \text{if} \quad b e < U\leq b e^2,\\
          2 - \ln(e^2-e+1) & \text{if} \quad U > b e^2.
        \end{array} \right.
    \end{align*}
\end{theorem}
We first note that Algorithm~\ref{alg:pred} is $1$-consistent, achieving the performance of the offline clairvoyant when the prediction is perfect.
The proof of Theorem \ref{theorem:cr_learn} in Section \ref{app:cr_learn} of the Supplementary Material details the competitive ratio as influenced by $\tau^*$, $L$, and $U$.\footnote{In Theorem 4.3, we present the competitive ratios for scenarios 1), 2), and, 3) separately, and for each scenario, we combine the results of the respective subroutines.}
Additionally, Section \ref{exp} demonstrates how varying the prediction 
confidence interval width $U-L$, while maintaining a constant $\tau^*$, affects performance. Notably, 
we observe that  Algorithm \ref{alg:pred} almost always outperforms its non-learning augmented equivalent across different interval widths.  
We defer the discussion of the design choice of $T$ relative to $b$ in the presence of prediction intervals to Remark~\ref{remark:design_choice_with_augmentation}.

\begin{remark}
\label{remark: th4.2}
Similarly, the term $\frac{1}{\tau^*}$ in the penalty can be replaced by a tuning parameter $\sigma$. In Section \ref{app:cr_learn}, we show that for $U \leq be^2$, the above results hold for a wide range of $\sigma$ values, specifically $\sigma \leq \frac{b}{e}$. However, when $T > be^2$, $\sigma$ should be of the order $\frac{1}{\tau^*}$to align the penalty term with the budget term in the objective.
\end{remark}

\begin{case}[t]
\caption{($i$, $b$, $\tilde{\tau}$, $L$, $U$, $\mathrm{Int}(\tilde \tau)$)}
\label{alg:aug_case1}
\begin{algorithmic}[1]
\IF{$i > \mathrm{Int}(\tilde{\tau}) + L$}
\STATE $\tilde{\tau} =  \tilde{\tau}e$
\ENDIF
\STATE $p_{i} = \frac{b}{\min(U, \tilde{\tau} + L)}$
\end{algorithmic}
\end{case}
\begin{case}[t]
\caption{($i$, $b$, $\tilde{\tau}$, $L$, $U$, $\mathrm{Int}(\tilde \tau)$)}
\label{alg:aug_case2}
\begin{algorithmic}[1]
\IF{$i > \mathrm{Int}(\tilde{\tau}) + L$}
\STATE $\tilde{\tau} =  \tilde{\tau}e$
\ENDIF
\STATE $p_{i} = \frac{b}{\min(U, \tilde{\tau} e + L )}$
\end{algorithmic}
\end{case}
\begin{case}[t]
\caption{($i$, $b$, $\tilde{\tau}$, $L$, $U$, $\mathrm{Int}(\tilde \tau)$)}
\label{alg:aug_case3}
\begin{algorithmic}[1]
\IF{$i > \mathrm{Int}(\tilde{\tau}) + L$}
\STATE $j = j+1$
\IF{$j = 2$}
\STATE $b = b(1-\frac{\tilde{\tau}+L-b}{\tilde{\tau}(e-1) + L}) $
\ELSE
\STATE $b = b(1-\frac{1}{e}) $
\ENDIF
\STATE $\tilde{\tau} =  \tilde{\tau}e$
\ENDIF
\IF{$j = 1$}
\STATE $p_{i} = \frac{b}{ \tilde{\tau}(e-1) +L}$
\ELSE
\STATE $p_{i} = \frac{b}{ \tilde{\tau}e}$
\ENDIF
\end{algorithmic}
\end{case}

\vspace{-5pt}
\section{Experiments}
\label{exp}

We numerically assess the performance of our proposed algorithms through  
both synthetic and real-world datasets. 

\vspace{-5pt}
\subsection{Synthetic Experiments}
%
%

\textbf{Benchmarks$\;$}
In the setting without learning augmentation, we compare Algorithm \ref{alg:rand}  against a conservative benchmark 
that delivers interventions with a constant probability $b / T$.
In the learning-augmented setting with
a confidence interval $[L,U]$, we
compare Algorithm \ref{alg:pred} against both a benchmark that delivers interventions with a constant probability $b / U$ and Algorithm \ref{alg:rand}. Due to limited algorithmic work on OUS and the absence of algorithms handling confidence intervals, we do not include additional benchmarks on the synthetic data. We also consider the SeqRTS algorithm \citep{liao2018a}, which ignores the prediction uncertainty of $\tau^*$, in the real-world example. 

\begin{figure*}[t!]
    \centering
    \begin{subfigure}[b]{0.25\linewidth}
        \centering
        \includegraphics[width=\linewidth,keepaspectratio]{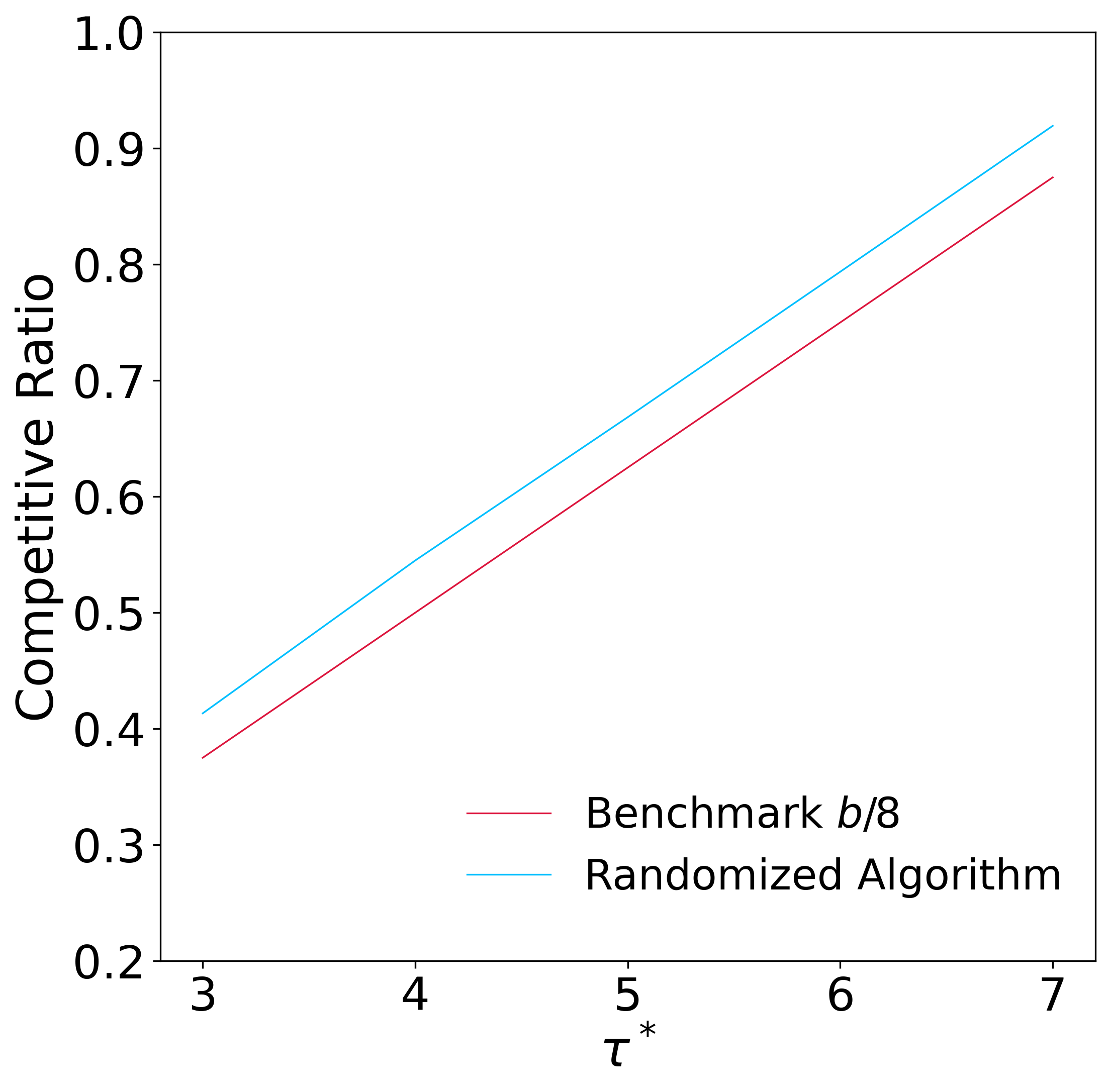}
        \vspace{-17pt}
        \caption{Scenario 1: $T=8$}
        \label{fig:set1a}
    \end{subfigure}
    \hfill
    \begin{subfigure}[b]{0.25\linewidth}
        \centering
        \includegraphics[width=\linewidth,keepaspectratio]{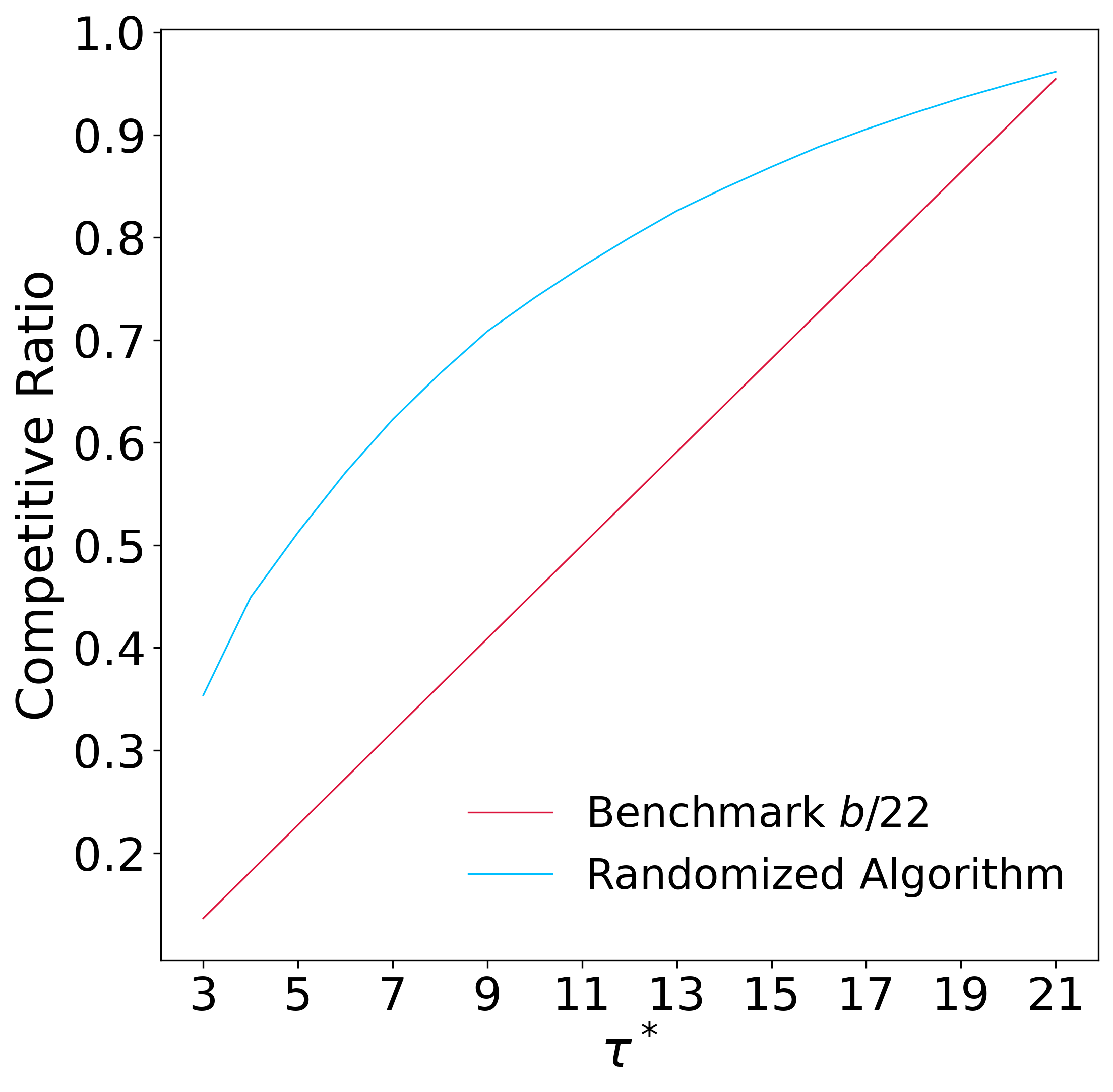}
        \vspace{-17pt}
        \caption{Scenario 2: $T=22$}
        \label{fig:set1b}
    \end{subfigure}
    \hfill
    \begin{subfigure}[b]{0.25\linewidth}
        \centering
        \includegraphics[width=\linewidth,keepaspectratio]{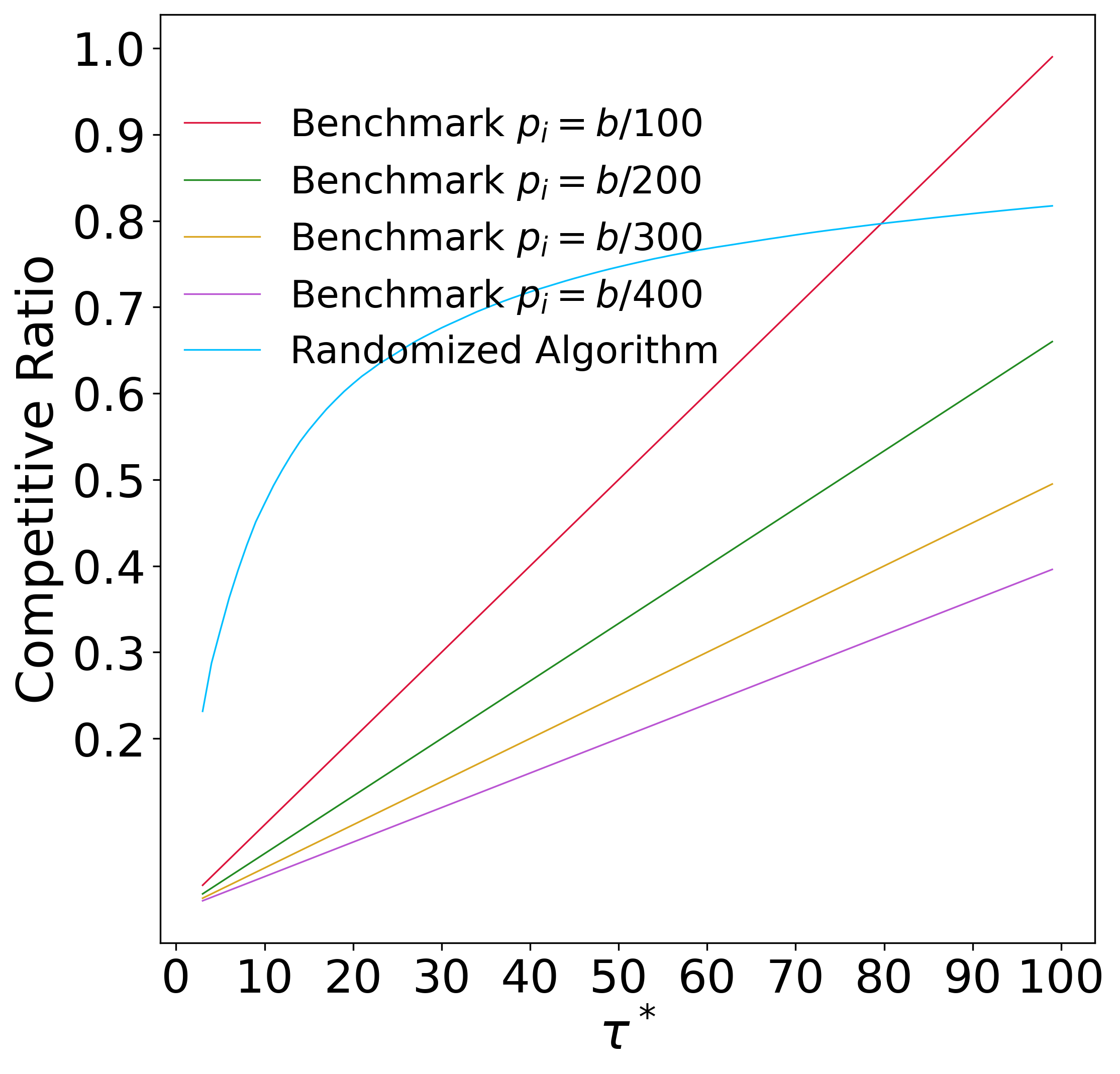}
        \vspace{-17pt}
        \caption{Scenario 3: $T=100$}
        \label{fig:set1c}
    \end{subfigure}
    \vspace{-8pt}
    \caption{Average competitive ratio under non-learning augmented setting with $b=3$. The scenarios correspond to $T\leq b e$, $be<T\leq be^2$, and $T>be^2$, respectively. }
    \label{fig:set1}
\end{figure*}

\textbf{Without Learning Augmentation$\;$}
In this 
setting, 
we test the performance of Algorithm~\ref{alg:rand} 
across all three scenarios listed in Theorem~\ref{theorem:cr_rand}.
We
fix the budget at $b=3$ and alter the horizon lengths $T$ to correspond with each case.
For Scenarios 1 and 2, we set $T$ to the maximum values allowed with $b=3$, specifically $T=8$ and $22$, as shown in Figure~\ref{fig:set1} (left and middle).  For Scenario 3, where $T$ can grow to infinity asymptotically, we set $T=100$ for simplicity (Figure~\ref{fig:set1} right). 
%
To simulate risk occurrences, 
we randomly draw an integer $\tau^*$ from the interval $[b, T-1]$ and randomly select
$\tau^*$ time points from $T$ as risk times.



Figure \ref{fig:set1} shows the average competitive ratio over a range of $\tau^*$ values. Figure~\ref{fig:set1a} , 
indicates that our randomized algorithm outperforms the benchmark by a constant competitive ratio across all values of $\tau^*$ in Scenario 1).
In Figure~\ref{fig:set1b},
our randomized algorithm increasingly outperforms the benchmark particularly as $\tau^*$ deviates away from the horizon length $T$ in Scenario 2). 
%
%
%
In Figure~\ref{fig:set1c}, as $T$ grows, the average competitive ratio of our algorithm would remain the same.\footnote{This is because when $b$ is fixed, the treatment assignment probability is independent of $T$.} 
Here, we conclude that our algorithm increasingly outperforms the benchmark as $T$ grows to infinity. 

\begin{remark}[\textbf{Design choice of $b$ and $T$ in absence of prediction confidence intervals}]
\label{remark:design_choice_T_no_augmentation}
 In real-world applications, the budget for intervention at each risk level is often fixed. Yet, there remains the design question of $T$, i.e., how granular to discretize the decision period. 
As shown in Figure~\ref{fig:set1}, while Scenario 3 offers the largest performance improvement as $T$ approaches infinity, our randomized algorithm achieves the highest competitive ratio across all $\tau^*$ in Scenarios 1 and 2. Therefore, in the absence of prediction intervals, we recommend setting $T$ below $b e^2$.
\end{remark}


\begin{figure*}[t!]
    \centering
    \begin{subfigure}[b]{0.25\linewidth}
        \centering
        \includegraphics[width=\linewidth,keepaspectratio]{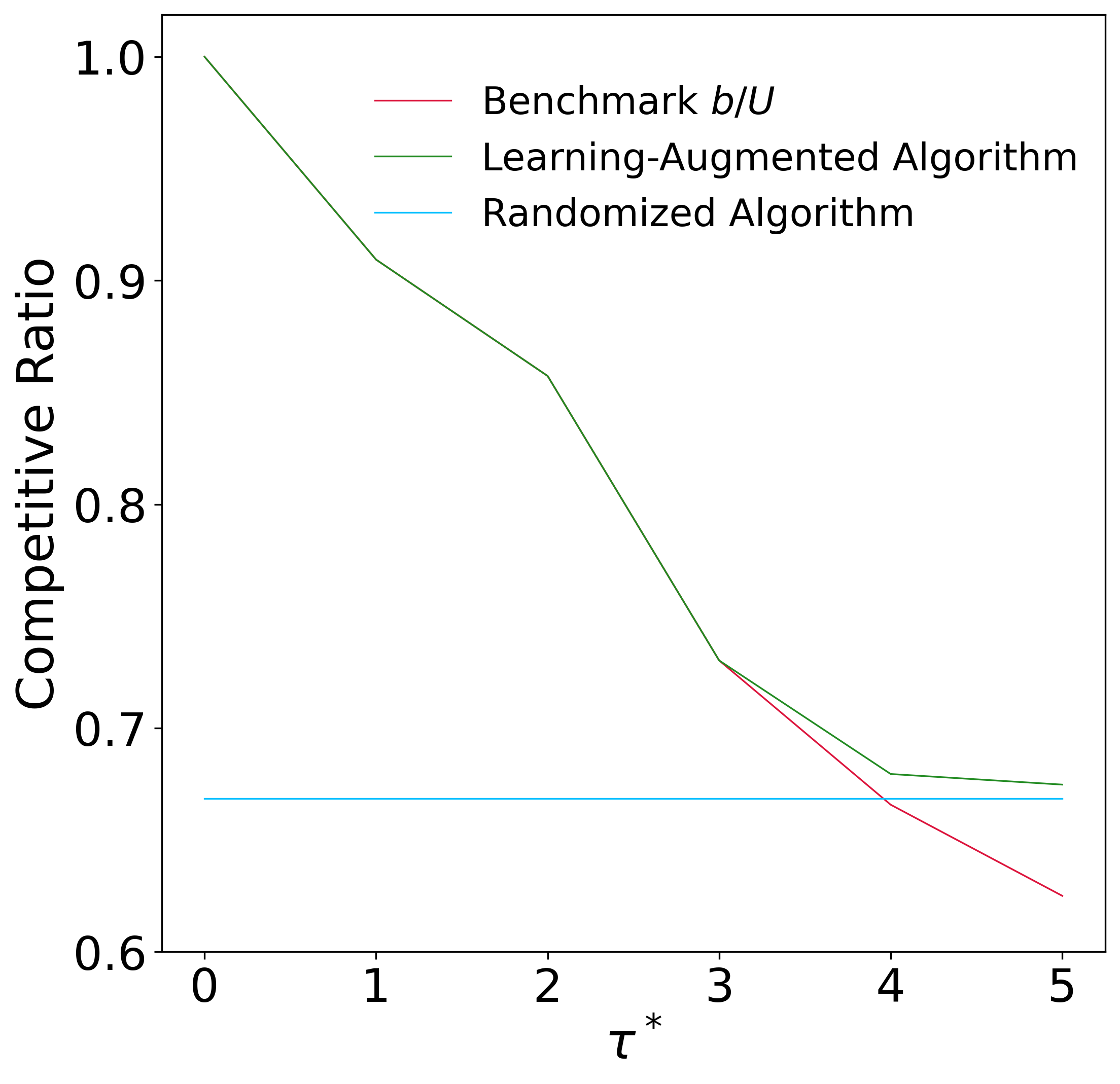}
        \vspace{-17pt}
        \caption{Scenario 1: $T=8$}
        \label{fig:set2a}
    \end{subfigure}
    \hfill
    \begin{subfigure}[b]{0.25\linewidth}
        \centering
        \includegraphics[width=\linewidth,keepaspectratio]{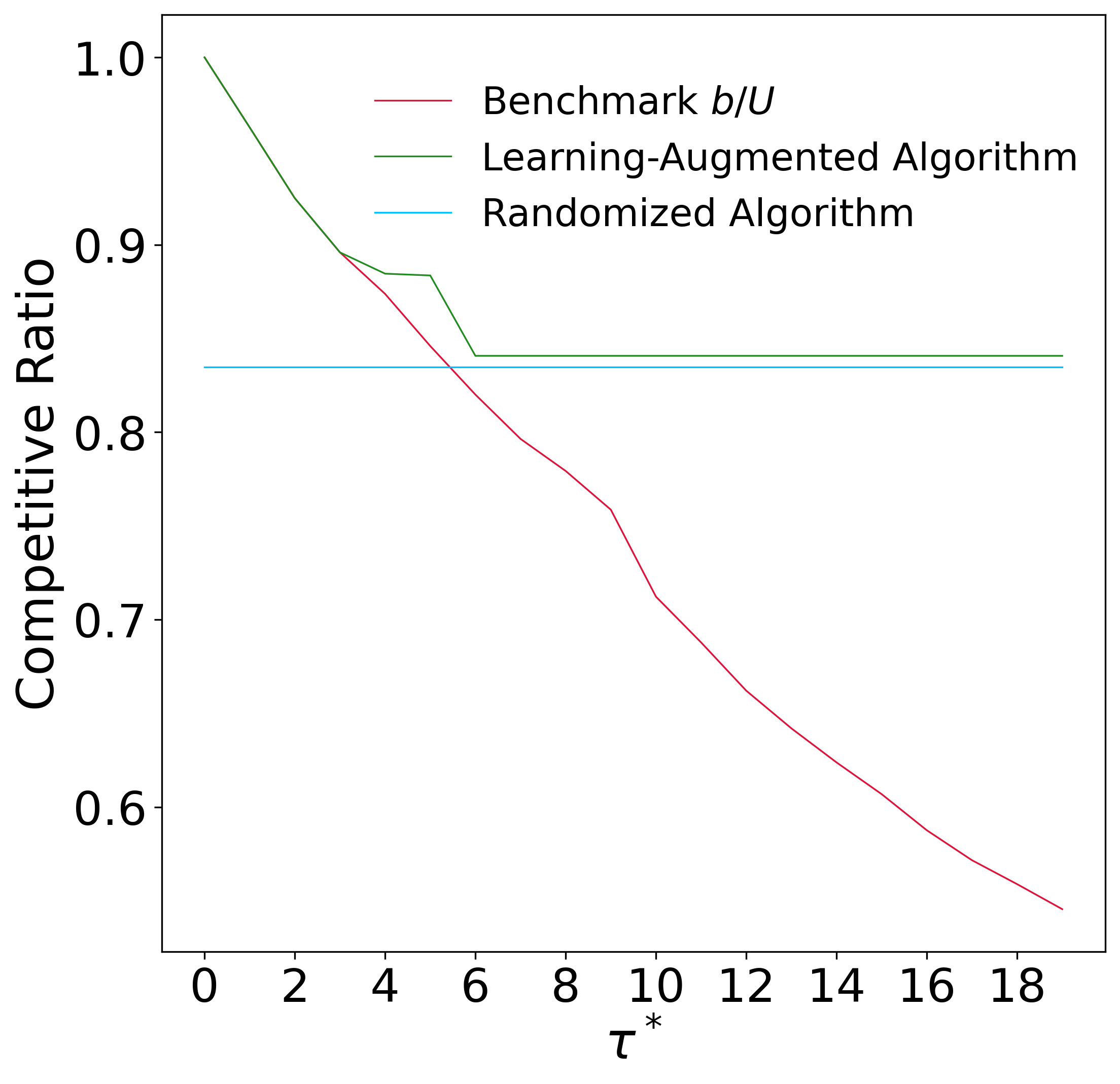}
        \vspace{-17pt}
        \caption{Scenario 2: $T=22$}
        \label{fig:set2b}
    \end{subfigure}
    \hfill
    \begin{subfigure}[b]{0.25\linewidth}
        \centering
        \includegraphics[width=\linewidth,keepaspectratio]{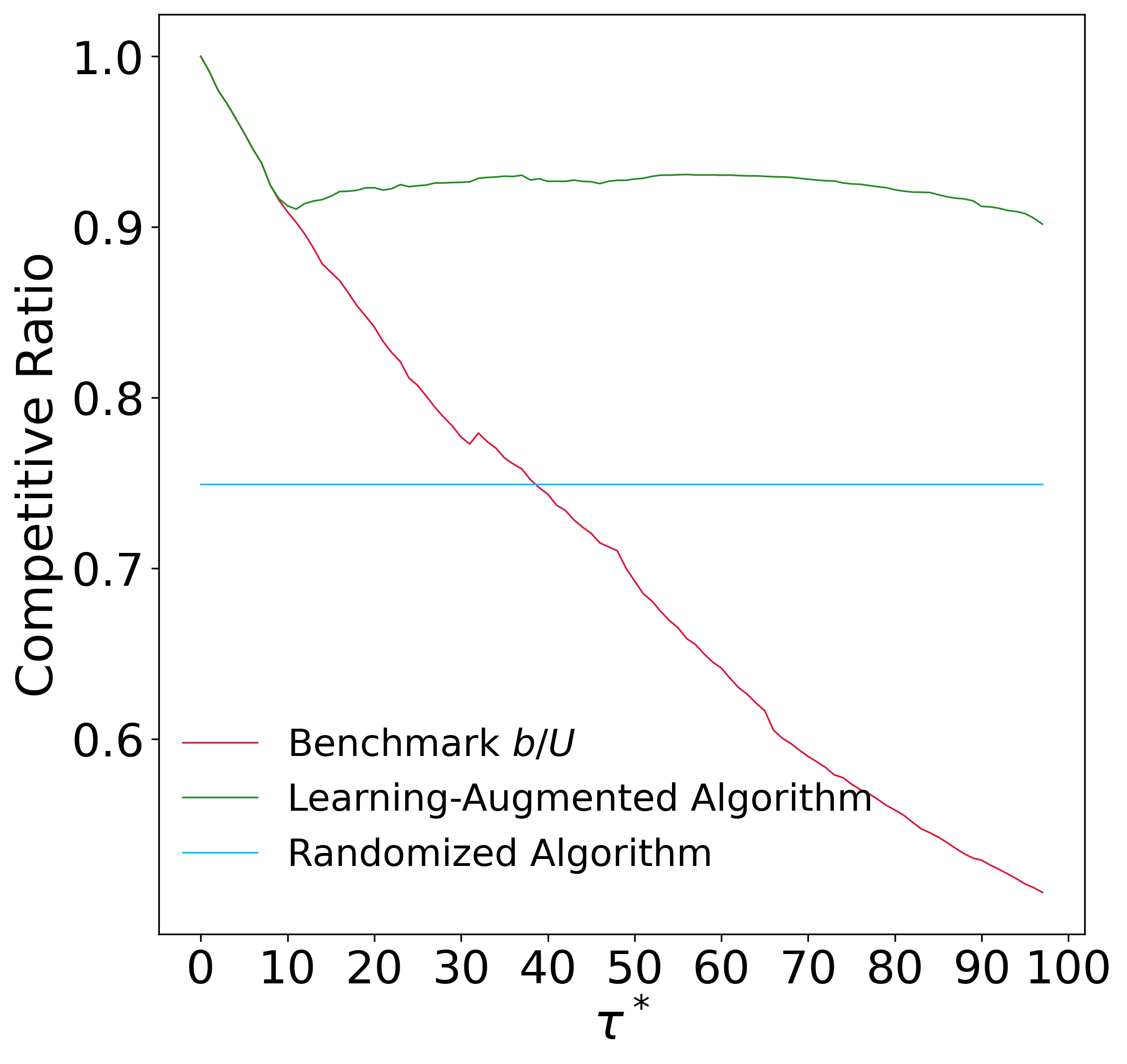}
        \vspace{-17pt}
        \caption{Scenario 3: $T=100$}
        \label{fig:set2c}
    \end{subfigure}
    \vspace{-8pt}
    \caption{Average competitive ratio under learning augmented setting with $b=3$. The scenarios correspond to $U\leq b e$, $be<U\leq be^2$, and $U>be^2$, respectively. }
    \label{fig:set2}
\end{figure*}

\textbf{With Learning Augmentation$\;$}
In this setting, we examine the performances of Algorithms \ref{alg:rand} and \ref{alg:pred} under varying prediction interval widths. Similar to the non-learning augmented setting, we keep the budget fixed at $b = 3$ and examine the performance of our learning-augmented algorithm under $T=8,22,$ and $100$, covering the three scenarios in Algorithm~\ref{alg:pred}. 
To compare the performance of our algorithm across various confidence widths, we fix 
$\tau^*= \text{Int}[0.5(T+b)]$, 
across all simulations shown in Figure~\ref{fig:set2}.\footnote{If we allow $\tau^*$ to change across different simulations, then the difference that we observe in competitive ratio might be due to this change in $\tau^*$.} 
The confidence intervals are randomly generated based on the given width and must contain $\tau^*$.

The average competitive ratio of each algorithm against a range of prediction interval widths is plotted in Figure \ref{fig:set2}. From Figure~\ref{fig:set2}, we observe that the naive benchmark (where $p_{i} = b/U$ for all $i\in[\tau^*]$) consistently outperforms the randomized algorithm (which does not have access to the prediction interval) when the confidence interval is narrow.  This is not surprising as in this case $\tau^*\approx U$. However, as the prediction interval widens, our randomized algorithm outperforms the naive benchmark. In addition, we observe that our learning-augmented algorithm performs no worse than both the naive benchmark and the randomized algorithm. In particular, the advantage of our algorithm is the largest when in Scenario 3. 

\begin{remark}[\textbf{Design choice of $b$ and $T$ in presence of prediction intervals}]
\label{remark:design_choice_with_augmentation}
If we expect the value of $\tau^*$ to be small, we recommend setting $T \leq be^2$ to ensure that the algorithm always operates in Scenario 2, where $U \leq be^2$. If we expect a reasonably large value of $\tau^*$, we recommend setting a large value for $T > be^2$ such that the algorithm operates under Scenario 3, where $U$ can exceed $be^2$. 
\end{remark}

Additional experimental results for small $\tau^*$ are provided in Section~\ref{app:results_smalltau} of the Supplementary Material.
We note that as $\tau^*$ decreases, the advantage of our algorithm in Scenario 2 increases. We also include competitive ratio figures without the penalization term from Problem~\eqref{raw_opt} in Section~\ref{app:results_budget}, measuring the fraction of the budget spent by our algorithms.

\vspace{-5pt}
\subsection{Real-World Experiments on HeartSteps}
Our research is inspired by the Heartsteps V1 mobile health study, which aimed to increase physical activity in 37 sedentary individuals over six weeks, with $T=144$ decision times per day \citep{klasnja2019efficacy}.
At each decision time $t$, we define the risk variable $R_t$ with a binary classification: $R_t=1$ indicates a sedentary state, identified by recording fewer than 150 steps in the prior 40 minutes, and $R_t = 0$ signifies a non-sedentary state. 
The primary objective of the OUS is to uniformly distribute around $b = 1.5$ interventions over sedentary times each day. 

\textbf{Benchmarks} In addition to the naive benchmark $b/U$, we evaluate Algorithms \ref{alg:rand} and \ref{alg:pred} against the SeqRTS algorithm, as introduced by \citet{liao2018a}. 
Under SeqRTS, 
the budget may be depleted before allocating for all available risk times. When this happens, a minimum probability of $1\times 10^{-6}$ is assigned to the remaining times when calculating the objective in Problem \eqref{raw_opt}. A detailed description of the SeqRTS method and further implementation details are included in Section \ref{app:heartsteps} of the Supplementary Material. The performances are measured by competitive ratio and entropy change averaged across user days. 

\begin{figure}[t!]
    \centering
\includegraphics[width=0.6\linewidth,keepaspectratio]{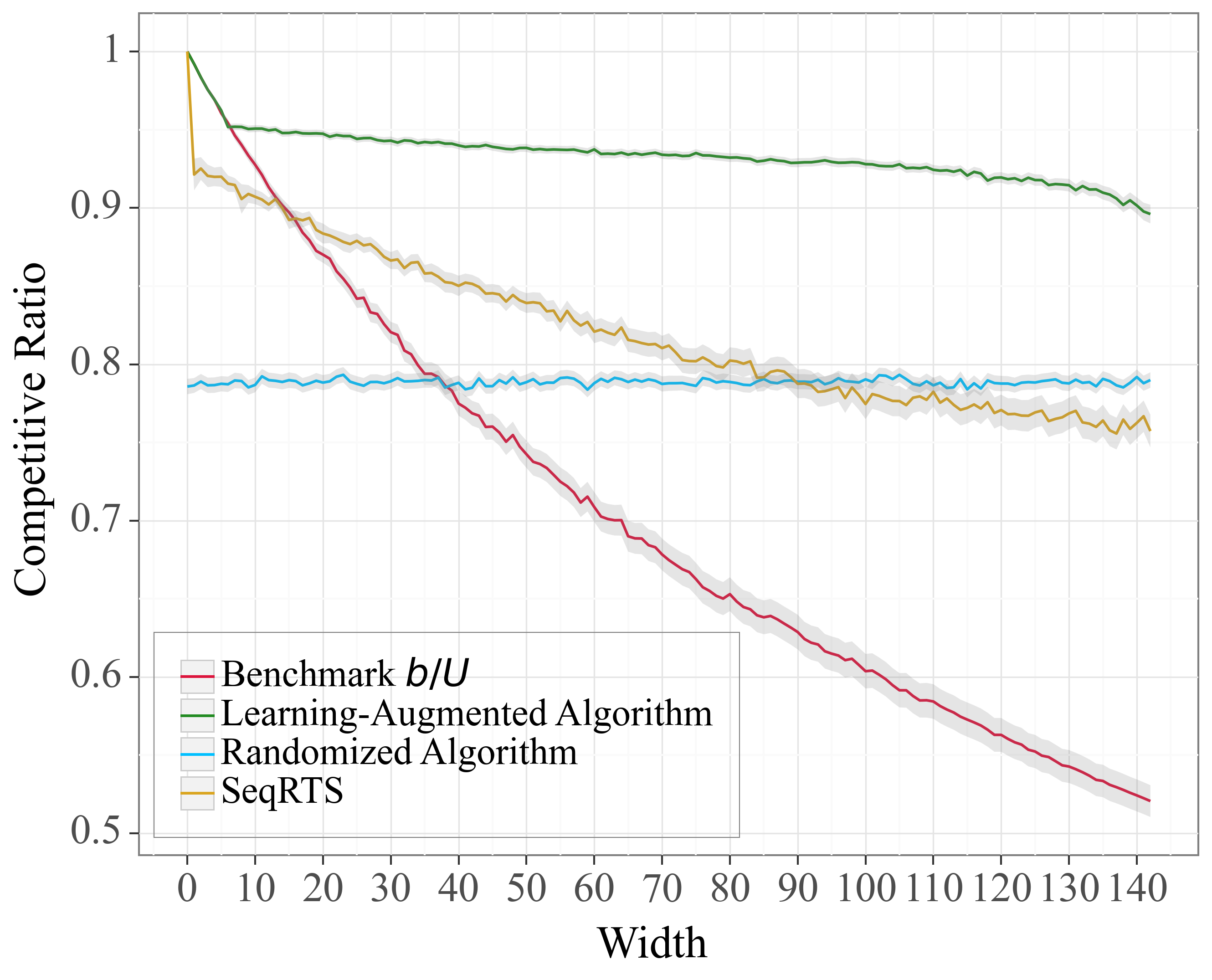}
        \vspace{-10pt}
    \caption{Average competitive ratio across user days under various prediction interval widths on HeartSteps V1 dataset. The shaded area indicates the $\pm 1.96$ standard error bounds across user days.}
    \vspace{-20pt}
    \label{fig:heart_a}
\end{figure}

In Figure \ref{fig:heart_a}, Algorithm \ref{alg:pred}, which incorporates a prediction interval, invariably outperforms the non-learning counterpart, the SeqRTS approach, and the naive benchmark $b/U$. Moreover, our algorithms exhibit superior uniformity in risk times sampling, evidenced by reduced entropy change compared to both the non-learning algorithm and SeqRTS, as further detailed in Figure \ref{fig:heart_b} in Section \ref{app:heartsteps} of the Supplementary Material.  
To better understand the behavior of SeqRTS, we set the minimum
probability to $0$ in Figure~\ref{fig:heart03} in Section~\ref{app:heartsteps}. 
This figure illustrates that SeqRTS could deplete its budget even when the prediction is fairly accurate, highlighting the robustness of our algorithms under adversarial risk level arrivals.

\textbf{Conclusion and Future Works$\;$}
This paper marks the first attempt to study the online uniform allocation problem within the framework of approximation algorithms. We introduce two novel online algorithms—either incorporating learning augmentation or not—backed by rigorous theoretical guarantees and empirical results. 
Future works include  
adapting existing algorithms to scenarios where prediction intervals improve over time.



\bibliography{citation}
\bibliographystyle{icml2024}


\appendix

\section{Extension to Multiple Risk Levels}
\label{app:mult}
In this section, we discuss the extension of the online uniform risk times sampling problem to multiple risk levels. 

At each time $t \in [1,T]$, the patient is associated with an ordinal risk level from $K$ possible levels.
The higher the risk level, the more likely the patient will experience a negative event, such as a relapse to smoking.
As stated previously, the distributions of risk levels are allowed to change \emph{arbitrarily} across decision periods since we anticipate that the treatment will reduce subsequent risk.

Let $\tau_k^*$ be the \emph{unknown} \emph{true} number of decision times at risk level $k \in [K]$ in a decision period, which is revealed at the end of the horizon $T$. 
For each risk level $k$, we define $p_{k,i_k}\in (0,1)$ to be the treatment probability at  time $i_k\in[\tau_k^*]$ .
The algorithm is provided with a \emph{soft} budget of $b_k$ for each risk level $k$, representing the total \emph{expected} number of interventions allowed to be delivered at risk level $k$ within each decision period. As before, we assume $\tau_k^*> b_k$ for technical convenience~\cite{liao2018a}.

Then
at each decision time $i_k$, 
the algorithm 
decides  
the intervention probability $p_{k, i_k}$. 
For each risk level $k$, the objectives of the online uniform allocation problem are to
1) assign the intervention probabilities $\{p_{k,i_k}\}_{i_k\in[\tau^*_k]}$  as uniform as possible across risk times, and 2) maximize the sum of intervention probabilities across risk times while adhering to the budget constraint $b_k$. 
 
For every risk level $k\in[K]$, we define the following optimization problem:



\vspace{-20pt}
\begin{align}\label{raw_opt_m}
\max \sum_{i_k}^{\tau_k^*} p_{k,i_k} &-  \frac{1}{\tau_k^*}\ln\left(\frac{\max_{i_k\in[\tau_k^*]}p_{k,i_k}}{\min_{i_k\in[\tau_k^*]}p_{k,i_k}}\right)\nonumber\\
    \text{s.t. } \mathbb{E}\Bigg[\sum_{i_k=1}^{\tau_k^*}&
    p_{k,i_k}\Bigg]  \leq b_k \nonumber \\
    p_{k,i_k}&\in (0,1) \quad \forall i\in[\tau_k^*].
\end{align}

Notably,  the proposed algorithms offer a feasible solution to the above optimization problem, allowing us to address each risk level independently.

\section{The Penalty Term for Uniformity}
\label{app:penalty}
We have previously considered statistical distance measures for quantifying the uniformity objective.  One important measure is the Kullback-Leibler (KL) divergence. However, this measure is not well defined in our setting since the optimal solution (which is a point mass on $b/\tau^*$) and the solutions given by our proposed algorithms are not defined on the same sample space. 

Recall that for two discrete distributions $P$ and $Q$ defined on the same sample space $\mathcal{X}$, the KL divergence is given by
\[
D_{KL}(P\|Q) = \sum_{x\in \mathcal{X}}P(x)\log\frac{P(x)}{Q(x)},
\]
where $P$ represents the data distribution, i.e., the optimal solution, and $Q$ represents an approximation of $P$, i.e., the solution given by an algorithm.

Let us consider a toy example where $\tau^* = b(e-1)$. In this case, the optimal solution should be $p_i = \frac{b}{b(e-1)} = \frac{1}{e-1}$ for each risk time $i\in[\tau^*]$. The corresponding distribution is a point mass, meaning the sample space $\mathcal{X}$ consists of a single element $(p_1=\frac{1}{e-1},\cdots, p_{\tau^*}=\frac{1}{e-1})$ with probability $1$. The solutions given by our proposed algorithms are of the form $(p_1,\cdots,p_{\tau^*})$, but the sample space $\mathcal{X}$ is $Q^{\tau^*}$, where the support of $Q$ is $(0,1)$. 

Clearly, the optimal solution and the solutions given by the proposed algorithms are not defined on the same sample space. Therefore, the KL divergence is not well-defined in this context.

\section{Proof for Algorithm 1}

\subsection{Proof of Lemma 3.1: Budget constraint} \label{app:budget_rand}
\begin{proof} We prove that the budget constraint is satisfied in expectation under each subroutine in Algorithm \ref{alg:rand}. 

\textbf{Subroutine 1}

Recall that $\tau^*$ is the true number of risk times. Here, we suppose $\tau^* = \beta e^{j^*}$ for some $j^* \in \mathbb{Z}^+$ and $\beta \in [  b,   be]$. Since $T \leq b e$, we have that $j^* = 0$.

In this analysis, our focus is solely on the worst-case scenario, where both $T$ and $\tau^*$ are very close to $b e$.  Assuming $\eta > b$ (where $\eta = \frac{T}{e-1}$), if this is not the case, the algorithm uniformly sets $p_{i} = \frac{b}{T}$ throughout.  

 When $\alpha$ falls in the range of $[b, \eta]$, Algorithm 1 starts with $p_{i} = \frac{b}{\alpha(e-1)}$ with a running length of $\alpha$, then transitions to $p_{i} = \frac{b}{T}$ for the second phase with a running length of $\beta - \alpha$. Otherwise, it consistently assigns $p_{i} = \frac{b}{T}$  with a running length of $\beta$. Therefore, the expected budget consists of two parts: one for $\alpha \in [b, \eta]$, where the term inside represents the used budget or the sum of treatment probabilities, and the other for $\alpha \in [\eta, b e]$, where the term inside represents the sum of treatment probabilities: 

\begin{align*}
 \mathbb{E}[\operatorname{Budget}]&= \int_{\eta}^{be} \frac{b}{T}\beta \frac{1}{\alpha}d\alpha + \int_{b}^{\eta} \Big[\frac{b}{\alpha(e-1)}\alpha + 
 \frac{b}{T}(\beta-\alpha)\Big] \frac{1}{\alpha}d\alpha \\
 &=\frac{b\beta}{T}\ln\frac{be}{\eta} + \frac{b}{e-1}\ln\frac{\eta}{b} + \frac{b\beta}{T}\ln\frac{\eta}{b} - \frac{b}{T}(\eta -   b)\\
 &=\frac{b\beta}{T} + \frac{b}{e-1}\ln\frac{\eta}{b} - \frac{b \eta}{T} + \frac{b^2}{T} \\
 &\leq b + \frac{b}{e-1}\ln\frac{T}{b(e-1)} - \frac{b}{e-1}  + \frac{b^2}{T}  \quad \text{increasing with} ~ \beta ~( \beta=T)\\
 &\leq b  -\frac{b}{e-1} + \frac{b}{e-1} - \frac{b}{e-1}\ln(e-1)+\frac{b}{e} \quad \text{increasing with} ~ T ~( T = be)\\
 & \leq  b - \frac{b}{e-1}\ln(e-1)+\frac{b}{e}\\
 &\approx  b.
\end{align*}
    

\textbf{Subroutine 2}

Reiterating our initial assumption, we set $\tau^*=\beta e^{j^*}$. The condition $b e < T \leq b e^2$ limits $j^*$ to either $0$ or $1$. However, our analysis is particularly concerned with the worst-case scenario, hence we consider only the case where $j^*=1$. 

When $\alpha$ falls in the range of $[b,\beta]$, Algorithm 1 starts with $p_i = \frac{b}{\alpha (e-1)}$ with a running length of $\alpha$, transitions to $p_i = \frac{b}{\alpha e (e-1)}$ with a running length of $\alpha e - \alpha$, and then continues with $p_i = \frac{b}{e^3}$ with a running length of $\beta e - \alpha e$. Otherwise, Algorithm 1 starts with $p_i = \frac{b}{\alpha (e-1)}$ with a running length of $\alpha$, transitions to $p_i = \frac{b}{\alpha e (e-1)}$ with a running length of $\beta e - \alpha$. 
The expected budget is therefore
\begin{align*}
    \mathbb{E} \left[\operatorname{Budget}\right] &= \int_{\beta}^{  b e} \left[\frac{b}{\alpha (e-1)}\alpha + \frac{b}{\alpha e (e-1)}(\beta e-\alpha)\right] \frac{1}{\alpha} d\alpha + \int_{  b}^\beta \Big[\frac{b}{\alpha (e-1)}\alpha\\
    &\quad+ \frac{b}{\alpha e (e-1)}(\alpha e -\alpha)+\frac{b}{\alpha e^3}(\beta e - \alpha e)\Big]\frac{1}{\alpha} d\alpha\\
    &=\frac{b}{e-1} + \frac{b\beta}{e-1}(\frac{1}{\beta}-\frac{1}{  be})-\frac{b}{e(e-1)}\ln\frac{  be}{\beta} + \frac{b}{e}\ln \frac{\beta}{  b}+\frac{b\beta}{e^2}(\frac{1}{  b}-\frac{1}{\beta})-\frac{b}{e^2}\ln\frac{\beta}{  b}\\
    &= \frac{b}{e-1} + \frac{b}{e-1} - \frac{\beta}{  e(e-1)} +\frac{\beta}{  e^2}-\frac{b}{e^2} - \frac{b}{e(e-1)} +(\frac{b}{e}+\frac{b}{e(e-1)}-\frac{b}{e^2})\ln\frac{\beta}{  b}\\
    &=\frac{b}{e-1} + \frac{b}{e} -\frac{b}{e^2} -\frac{\beta}{  e^2(e-1)} + (\frac{b}{e-1}-\frac{b}{e^2})\ln \frac{\beta}{  b}\\
    &\leq \frac{b}{e-1} + \frac{b}{e} -\frac{b}{e^2} -\frac{  be}{  e^2(e-1)} + (\frac{b}{e-1}-\frac{b}{e^2})\ln \frac{  be}{  b}\quad \text{increasing with} ~ \beta ~( \beta=  be)\\
    &=\frac{2b}{e}+\frac{b}{e-1}-\frac{2b}{e^2} \approx b
\end{align*}

\textbf{Subroutine 3}

Under the assumption of $\tau^* = \beta e^{j^*}$, and given the condition $T > b e^2$, it is possible for $j^*$ to be $0$ or to extend towards infinity. Our focus, however, is confined to the worst-case scenarios, particularly those where $j^* \geq 1$.

When $\alpha$ falls in the range of $[b, \beta]$. Algorithm 1 stops with $p_i=\frac{b(1-1/e)^{j^*}}{\alpha e^{j^*+2}}$ with a running length of $\beta e^{j^*} - \alpha e^{j^*}$. Otherwise, Algorithm 1 stops with $p_i=\frac{b(1-1/e)^{j^*-1}}{\alpha e^{j^*+1}}$ with a running length of $\beta e^{j^*} - \alpha e^{j^*-1}$. Therefore, the expected budget is
\begin{align*}
    \mathbb{E} &\left[\operatorname{Budget}\right] = \int_{\beta}^{  be}\left[\frac{b}{\alpha e} \alpha + \sum_{i=2}^{j^*} \frac{b (1-1/e)^{j-2}}{\alpha e^j}(\alpha e^{j-1} - \alpha e^{j-2}) + \frac{b (1-1/e)^{j^*-1}}{\alpha e^{j^*+1}}(\beta e^{j^*} -\alpha e^{j^*-1})\right] \frac{1}{\alpha}d \alpha \\
    &\quad + \int_{  b}^{\beta} \left[\frac{b}{\alpha e}\alpha + \sum_{j=2}^{j^*+1}\frac{b(1-1/e)^{j-2}}{\alpha e^j}(\alpha e^{j-1}-\alpha e^{j-2}) + \frac{b (1-1/e)^{j^*}}{\alpha e^{j^*+2}}(\beta e^{j^*}-\alpha e^{j^*})\right]\frac{1}{\alpha}d \alpha\\
    & = \int_{\beta}^{  be}\left[\frac{b}{e} +b (1-\frac{1}{e}-(1-\frac{1}{e})^{j^*}) + \frac{b(e-1)^{j^*-1}}{e^{j^*+1}}\frac{\beta e-\alpha}{\alpha} \right]\frac{1}{\alpha}d \alpha\\
    &\quad + \int_{  b}^{\beta} \left[\frac{b}{e} + b(1-\frac{1}{e}-(1-\frac{1}{e})^{j^*+1}) + \frac{b (e-1)^{j^*}}{e^{j^*+2}}\frac{\beta-\alpha}{\alpha}\right]\frac{1}{\alpha}d \alpha\\
    & = \frac{b}{e} + b(1-\frac{1}{e}) - b(1-\frac{1}{e})^{j^*}\ln \frac{  be}{\beta}- b(1-\frac{1}{e})^{j^*+1}\ln \frac{\beta}{  b}\\
    &\quad + \frac{b(e-1)^{j^*-1}}{e^{j^*+1}} \left[e-\frac{\beta}{  b}+\ln \frac{\beta}{  b e}\right] + \frac{b(e-1)^{j^*}}{e^{j^*+2}}\left[\frac{\beta}{  b}-1+\ln\frac{  b}{\beta}\right]\\
    &\leq b- b(1-\frac{1}{e})^{j^*+1}\ln \frac{  b e}{  b} + \frac{b(e-1)^{j^*}}{e^{j^*+2}}\left[\frac{  b e}{  b}-1+\ln\frac{  b}{  b e}\right]\quad \text{increasing with} ~ \beta ~( \beta=  be)\\
    &= b -b(1-\frac{1}{e})^{j^*+1} + \frac{b(e-1)^{j^*}(e-2)}{e^{j^*+2}} \leq b
\end{align*}

By combining the above results, we establish Lemma \ref{lemma:feasibility_rand}.
\end{proof}

\subsection{Proof of Theorem 3.1: Competitive Ratio}  
\label{app:cr_rand}
\begin{proof} In what follows we derive the competitive ratio under each subroutine.

\textbf{Subroutine 1}

Recall that $\tau^*$ represents the true number of available risk times at risk level $k$, we assume $\tau^* = \beta e^{j^*}$, where $j^* \in \mathbb{Z}^+$ and $\beta \in [b, be]$. It's evident that when $T \leq b e$, $j^* = 0$ follows naturally.
    
Define $\eta = T/(e-1)$. Let us first consider the case where $\eta \leq b$, leading to $T\leq b(e-1)$. Given that $p_{ i} = \frac{b}{\min(T, \tilde \tau (e-1))}$, it follows that the algorithm consistently sets $p_{i} = \frac{b}{T}$. Consequently, we have
\begin{align*}
     \mathbb{E}[\operatorname{SOL}]&= \frac{b}{T}\beta \geq \frac{b}{  b(e-1)}\beta \geq \frac{b}{e-1}.
\end{align*}

Next, let us consider the case where $\eta > b$. We focus on two cases: (1) $\beta < \eta$ and (2) $\beta \geq \eta$.

Suppose $\beta < \eta$. When $\alpha$ falls within $[b, \beta]$, the algorithm initiates with $p_{i} = \frac{b}{\alpha(e-1)}$ with a running length of $\alpha$, then adjusts to $p_{i} = \frac{b}{T}$ in the subsequent round with a running length of $\beta - \alpha$; when $\alpha$ falls in the range of $[\beta, \eta]$, the algorithm initiates with $p_{i} = \frac{b}{\alpha(e-1)}$ with a running length of $\beta$ and stops on this stage; otherwise, it consistently uses $p_{i} = \frac{b}{T}$ with a running length of $\beta$. Therefore, the expected solution is


\begin{align*}
    \mathbb{E}[\operatorname{SOL}]&= \int_{\eta}^{  be} \frac{b}{T}\beta \frac{1}{\alpha} d\alpha + \int_{  b}^{\beta} \Big[\frac{b}{\alpha (e-1)}\alpha + \frac{b}{T}(\beta - \alpha) - \sigma \ln \frac{T}{\alpha(e-1)} \Big] \frac{1}{\alpha}d\alpha \\
    & \quad + \int_{\beta}^{\eta}\frac{b}{\alpha(e-1)}\beta \frac{1}{\alpha}d\alpha\\
    &= \frac{b\beta}{T}\ln\frac{  be}{\eta}+\frac{b}{e-1}\ln \frac{\beta}{  b} + \frac{b\beta}{T}\ln \frac{\beta}{b} - \frac{b}{T}(\beta-  b)+ \frac{\sigma}{2}(\ln(\frac{T}{\beta (e-1)})^2 - \ln(\frac{T}{b (e-1)})^2) \\
    &\quad +\frac{b\beta}{e-1}(\frac{1}{\beta} - \frac{1}{\eta})\\
    &\geq \frac{  b^2}{T}\ln\frac{  be(e-1)}{T} + \frac{b}{e-1} - \frac{  b^2}{T} \quad \text{increasing with} ~ \beta ~( \beta=  b)\\
    &\geq \frac{b}{e}\ln (e-1) + \frac{b}{e(e-1)} \quad \text{decreasing with} ~ T ~( T=   be).
\end{align*}

Suppose $\beta \geq \eta$.  It follows that the algorithm always proceeds to the second round. When $\alpha$ falls within $[b, \eta]$, the algorithm initiates with $p_{i} = \frac{b}{\alpha(e-1)}$ with a running length of $\alpha$, then adjusts to $p_{i} = \frac{b}{T}$ in the subsequent round with a running length of $\beta - \alpha$; otherwise, it consistently uses $p_{i} = \frac{b}{T}$ with a running length of $\beta$. 
Consequently, we have

\begin{align*}
 \mathbb{E}[\operatorname{SOL}]&= \int_{\eta}^{be} \frac{b}{T}\beta \frac{1}{\alpha}d\alpha + \int_{  b}^{\eta} \Big[\frac{b}{\alpha(e-1)}\alpha + 
 \frac{b}{T}(\beta-\alpha)- \sigma\ln \frac{T}{\alpha (e-1)} \Big] \frac{1}{\alpha}d\alpha \\
 &=\frac{b\beta}{T}\ln\frac{  be}{\eta} + \frac{b}{e-1}\ln\frac{\eta}{  b} + \frac{b\beta}{T}\ln\frac{\eta}{  b} - \frac{b}{T}(\eta -   b) + \frac{\sigma}{2}(\ln(\frac{T}{\eta (e-1)})^2 - \ln(\frac{T}{b (e-1)})^2) \\
 &\geq \frac{b}{e-1}\ln\frac{  be}{\eta} + \frac{2b}{e-1}\ln\frac{T}{b(e-1)} - \frac{b}{e-1}+\frac{  b^2}{T} - \frac{\sigma}{2} \ln (\frac{T}{b(e-1)})^2\\ &\quad \text{increasing with} ~ \beta ~( \beta=\eta= T/(e-1))\\
 &\geq \frac{2b}{e-1}-\frac{b}{e-1}\ln(e-1) -\frac{b}{e(e-1)} - \frac{\sigma}{2}\ln(\frac{e}{e-1})^2 \quad \text{decreasing with} ~ T ~( T=  be).
\end{align*}

\textbf{Subroutine 2}

Recall that for $b e < T \leq b e^2$, $j^*$ is restricted to being either $0$ or $1$. Below we separately consider these two cases. 
 
Suppose $j^* = 0$. When $\alpha$ falls in the range of $[b,\beta]$, Algorithm 1 begins with $p_i = \frac{b}{\alpha(e-1)}$ with a running length of $\alpha$, then transitions to $p_i = \frac{b}{\alpha e (e-1)}$ with a running length of $\beta - \alpha$. Otherwise, Algorithm 1 begins with $p_i = \frac{b}{\alpha(e-1)}$ with a running length of $\beta$ and stops. 
It follows that
\begin{align*}
 \mathbb{E}[\operatorname{SOL}]&=\int_\beta^{  b e} \frac{b}{\alpha(e-1)}   \beta   \frac{1}{\alpha}   d \alpha+\int_{  b}^\beta\left[\frac{b}{\alpha(e-1)}   \alpha+\frac{b}{\alpha   e  (e-1)}  (\beta-\alpha)-\sigma \ln (e) \right]   \frac{1}{\alpha} d \alpha\\
    &= \frac{b \beta}{e-1}  \left(\frac{1}{\beta}-\frac{1}{  b e}\right)+\frac{b}{e-1}  \left(\ln \beta-\ln   b\right)+\frac{b \beta}{e(e-1)}  \left(\frac{1}{  b}-\frac{1}{\beta}\right)\\
    &\quad -\frac{b}{e(e-1)}\left(\ln \beta-\ln   b\right) - \sigma \ln \frac{\beta}{b}\\
    &=\frac{b}{e} + \frac{b}{e}\ln \frac{\beta}{b} - \sigma \ln\frac{\beta}{b}\\
    &\geq \frac{b}{e} \quad \text{increasing with} ~ \beta ~( \beta=  b).
\end{align*}

Suppose $j^* = 1$.  When $\alpha$ falls in the range of $[b,\beta]$, Algorithm 1 begins with $p_i = \frac{b}{\alpha(e-1)}$ with a running length of $\alpha$,  transitions to $p_i = \frac{b}{\alpha e (e-1)}$ with a running length of $\alpha e - \alpha$, and then continues with $p_i = \frac{b}{e^3}$ with a running length of $\beta e - \alpha e$. Otherwise, Algorithm 1 begins with $p_i = \frac{b}{\alpha(e-1)}$ with a running length of $\alpha$, then transitions to $p_i = \frac{b}{\alpha e (e-1)}$ with a running length of $\beta e -\alpha$. 
Therefore, the expected solution is
 \begin{align*}
 \mathbb{E}[\operatorname{SOL}]&=\int_\beta^{  b e}\left[\frac{b}{\alpha  (e-1)}   \alpha+\frac{b}{\alpha   e(e-1)}  (\beta e-\alpha) - \sigma \ln e \right]   \frac{1}{\alpha} d \alpha\\
     &\quad +\int_{  b}^\beta\left[\frac{b}{\alpha(e-1)}   \alpha+\frac{b}{\alpha   e  (e-1)}  (\alpha e-\alpha)+\frac{b}{\alpha   e^3}  (\beta e-\alpha e) - \sigma \ln \frac{e^3}{e-1}\right]   \frac{1}{\alpha} d \alpha\\
     &=\frac{b}{e-1} \ln \frac{b e}{\beta} +\frac{b \beta}{e-1}  \left(\frac{1}{\beta}-\frac{1}{  b e}\right)-\frac{b}{e(e-1)}  \ln \frac{b e}{\beta} -\sigma \ln \frac{b e}{\beta}\\
     &\quad +\frac{b}{e-1} \ln \frac{\beta}{b}  +\frac{b}{e}\ln \frac{\beta}{b} +\frac{b \beta}{e^2}\left(\frac{1}{  b}-\frac{1}{\beta}\right)-\frac{b}{e^2} \ln \frac{\beta}{b} - \sigma \ln \frac{e^3}{e-1}\ln \frac{\beta}{b}\\
    &= \frac{b}{e}\ln \frac{b e}{\beta} + \frac{b}{e-1} - \frac{\beta}{e(e-1)} - \sigma\ln \frac{b e}{\beta}\\
    &\quad + \frac{b}{e-1}\ln \frac{\beta}{b} + \frac{b}{e}\ln \frac{\beta}{b} + \frac{\beta}{e^2} - \frac{b}{e^2} - \frac{b}{e^2}\ln \frac{\beta}{b} - \sigma \ln \frac{e^3}{e-1}\ln \frac{\beta}{b}\\
    &\geq \frac{b}{e} + \frac{b}{e-1} -\frac{b}{e(e-1)} - \sigma \quad \text{increasing with} ~ \beta ~( \beta=   b) \\
    &= \frac{2b}{e}- \sigma.
 \end{align*}

\textbf{Subroutine 3}

In the scenarios where $T > b e^2$, we consider two cases: (1) $j^* \geq 1$ and (2) $j^* = 0$. 

Let us first consider the case where $j^* \geq 1$. If $\alpha \geq \beta$, the algorithm stops at the $j^*+1$th round by design of the algorithm ($\alpha e^{j^*} \geq \beta e^{j^*}$); on the other hand, if $\alpha < \beta$, the algorithm stops at the $j^*+2$th round ($\alpha e^{j^*+1} \geq \beta e^{j^*}$). 
The objective function when $\alpha \geq \beta$ is
    \begin{align*}
       \text{SOL}_{ 1} &= \sum_{j=1}^{j^*} \frac{b\left(1-\frac{1}{e}\right)^{j-2}}{\alpha  e^{j} } \left(\alpha e^{j-1}-\alpha e^{j-2}\right)+\frac{ b\left(1-\frac{1}{e}\right)^{j^*-1}}{\alpha  e^{j^*+1}} \left(\beta e^{j^*}-\alpha e^{j^*-1}\right) - \sigma \ln \frac{e^{2j^*-1}}{(e-1)^{j^*-1}}\\
       &= \sum_{j=1}^{j^*} \frac{b(e-1)^{j-1}}{e^j}+\frac{b(e-1)^{j^*-1}}{e^{j^*+1}}  \frac{\beta e-\alpha}{\alpha}-\sigma \ln \frac{e^{2j^*-1}}{(e-1)^{j^*-1}}\\
       &= b\left(1-\left(1-\frac{1}{e}\right)^{j^*}\right)+\frac{b(e-1)^{j^*-1}}{e^{j^*+1}}  \frac{\beta e-\alpha}{\alpha} - \sigma \ln \frac{e^{2j^*-1}}{(e-1)^{j^*-1}}.
    \end{align*}

    The objective function when $\alpha < \beta$ is \
    \begin{align*}
        \text{SOL}_{2} &= \sum_{j=1}^{j^*+1} \frac{b\left(1-\frac{1}{e}\right)^{j-2}}{\alpha  e^{j}} \left(\alpha e^{j-1}-\alpha  e^{j-2}\right)+\frac{b\left(1-\frac{1}{e}\right)^{j^*}}{\alpha  e^{j^*+2}} \left(\beta  e^{j *}-\alpha e^{j *}\right) - \sigma \ln \frac{e^{2j^*+1}}{(e-1)^{j^*}}\\
        &=\sum_{j=1}^{j^*+1} \frac{b(e-1)^{j-1}}{e^j}+\frac{b(e-1)^{j *}}{e^{j^*+2}} \frac{\beta-\alpha}{\alpha}- \sigma \ln \frac{e^{2j^*+1}}{(e-1)^{j^*}}\\
        &=b\left(1-\left(1-\frac{1}{e}\right)^{j^*+1}\right) +\frac{b(e-1)^{j^*}}{e^{j^*+2}}  \frac{\beta-\alpha}{\alpha}- \sigma \ln \frac{e^{2j^*+1}}{(e-1)^{j^*}}.
    \end{align*}

    The expected value of our solution is
    \begin{equation}
    \mathbb{E}[\operatorname{SOL}]=\int_\beta^{  be} \operatorname{SOL}_{1} f(\alpha) d \alpha+\int_{b}^\beta \operatorname{SOL}_{2} f(\alpha) d \alpha.
    \end{equation}

    Notice that
    \begin{align*}
        \int_{\beta}^{  b e} \operatorname{SOL}_{1} f(\alpha) d \alpha &=b\left(1-\left(1-\frac{1}{e}\right)^{j^*}\right) \int_\beta^{  b e} \frac{1}{\alpha} d \alpha + \frac{b(e-1)^{j^*-1}}{e^{j^*+1}} \int_\beta^{  b e} \frac{\beta e-\alpha}{\alpha}  \frac{1}{\alpha} d \alpha\\
        & - \left[ \sigma \ln \frac{e^{2j^*-1}}{(e-1)^{j^*-1}}\right]  \int_{\beta}^{b e}\frac{1}{\alpha} d \alpha\\
        &= b\left(1-\left(1-\frac{1}{e}\right)^{j^*}\right)\ln \frac{  be}{\beta}+ \frac{b(e-1)^{j^*-1}}{e^{j^*+1}}\left(e-\frac{\beta}{  b}-\ln \frac{b e}{\beta}\right)\\
        &-\left[ \sigma \ln \frac{e^{2j^*-1}}{(e-1)^{j^*-1}}\right] \ln \frac{b e}{\beta}.
    \end{align*}
    and
    \begin{align*}
        \int_{  b}^\beta \operatorname{SOL}_{2} f(\alpha) d \alpha &=b\left(1-\left(1-\frac{1}{e}\right)^{j^*+1}\right) \int_{  b}^\beta \frac{1}{\alpha} d \alpha+ \frac{b(e-1)^{j^*}}{e^{j^*+2}}  \int_{  b}^\beta \frac{\beta-\alpha}{\alpha}  \frac{1}{\alpha} d \alpha\\
        &- \left[ \sigma \ln \frac{e^{2j^*+1}}{(e-1)^{j^*}}\right]\int_{ b}^\beta \frac{1}{\alpha}    d \alpha\\
        &= b\left(1-\left(1-\frac{1}{e}\right)^{j^*+1}\right)\ln \frac{\beta}{  b}+\frac{b(e-1)^{j^*}}{e^{j *+2}}\left(\frac{\beta}{  b}-1-\ln \frac{\beta}{  b}\right)\\
        &- \left[\sigma \ln \frac{e^{2j^*+1}}{(e-1)^{j^*}}\right]\ln \frac{\beta}{b}.
    \end{align*}
Hence,
\begin{align*}
    \mathbb{E}&[\operatorname{SOL}] = b\left(1-\left(1-\frac{1}{e}\right)^{j^*}\right)\ln \frac{  be}{\beta} + b\left(1-\left(1-\frac{1}{e}\right)^{j^*+1}\right)\ln \frac{\beta}{  b}\\
    &\quad+  \frac{b(e-1)^{j^*-1}}{e^{j^*+1}}\left(e-\frac{\beta}{  b}-\ln \frac{  b e}{\beta}\right) + \frac{b(e-1)^{j^*}}{e^{j *+2}}\left(\frac{\beta}{  b}-1 - \ln \frac{\beta}{  b}\right)\\
    &\quad-\left[ \sigma \ln \frac{e^{2j^*-1}}{(e-1)^{j^*-1}}\right] \ln \frac{b e}{\beta} - \left[\sigma \ln \frac{e^{2j^*+1}}{(e-1)^{j^*}}\right]\ln \frac{\beta}{b}\\
    &= b - b\left(1-\frac{1}{e}\right)^{j^*}\left[\ln \frac{  b e}{\beta}+\left(1-\frac{1}{e}\right) \ln \frac{\beta}{  b} \right]\\
    & \quad+ \frac{b(e-1)^{j^*-1}}{e^{j^*+1}} \left[e-\frac{\beta}{  b}-\ln \frac{  be }{\beta}+\frac{e-1}{e}\left(\frac{\beta}{  b}-1-\ln \frac{\beta}{  b}\right)\right]\\
    &\quad -\left[ \sigma \ln \frac{e^{2j^*-1}}{(e-1)^{j^*-1}}\right] \ln \frac{b e}{\beta} - \left[\sigma \ln \frac{e^{2j^*+1}}{(e-1)^{j^*}}\right]\ln \frac{\beta}{b}\\
    &\geq b - b\left(1-\frac{1}{e}\right)^{j^*}+ b \left(1-\frac{1}{e}\right)^{j^*} \frac{e-2}{e(e-1)} - \sigma \ln \frac{e^{2j^*-1}}{(e-1)^{j^*-1}}\quad \text{increasing with} ~ \beta ~( \beta=  b).
\end{align*}

Now consider the case where $j^* = 0$. When $\alpha$ falls within $[b,\beta]$, Algorithm 1 starts with $p_i = \frac{b}{\alpha e}$ with a running length of $\alpha$, then transitions to $p_i = \frac{b}{\alpha e^2}$ with a running length of $\beta  - \alpha$. Otherwise, Algorithm 1 keeps $p_i = \frac{b}{\alpha e}$ for $\beta$ time points.
It follows that 
\begin{align*}
    \mathbb{E}[\operatorname{SOL}] &= \int_{\beta}^{b e}\frac{b}{\alpha e} \beta \frac{1}{\alpha} d\alpha + \int_{b}^{\beta}\left[\frac{b}{\alpha e}\alpha + \frac{b}{\alpha e^2}(\beta-\alpha) - \sigma \ln e\right]\frac{1}{\alpha}d\alpha\\
    &= \frac{b \beta}{e}(\frac{1}{\beta} - \frac{1}{b e}) + \frac{b}{e}\ln \frac{\beta}{b} + \frac{b \beta}{e^2}(\frac{1}{b}-\frac{1}{\beta}) - \frac{b}{e^2}\ln \frac{\beta}{b} -\sigma \ln \frac{\beta}{b}\\
    &= \frac{b}{e} -\frac{\beta}{e^2} + \frac{\beta}{e^2}-\frac{b}{e^2} + (\frac{b}{e} -\frac{b}{e^2})\ln \frac{\beta}{b}-\sigma \ln \frac{\beta}{b}\\
    &\geq b\left(\frac{1}{e}-\frac{1}{e^2}\right)\quad \text{increasing with} ~ \beta ~( \beta=b).
\end{align*}

\paragraph{Tuning parameter selection}
For Scenario 1) where $T \leq be$, the competitive ratio is the   
\begin{align*}
\min \left(\frac{1}{e}\left(\ln(e-1) + \frac{1}{e-1}\right), \frac{2}{e-1}-\frac{1}{e-1}\ln(e-1)-\frac{1}{e(e-1)}-\frac{\sigma}{b}(1-\ln (e-1)\right).    
\end{align*}

For Scenario 2) where $be < T \leq be^2$, the competitive ratio  is
\begin{align*}
    \min \left(\frac{1}{e}, \frac{2}{e} - \frac{\sigma}{b} \right).
\end{align*}

For Scenario 3) where $T > be^2$, the competitive ratio is
\begin{align*}
    \min \Bigg(\frac{1}{e}-\frac{1}{e^2}, 1-(1-\frac{1}{e})^{j^*}+(1-\frac{1}{e})^{j^*}\frac{e-2}{e(e-1)}-\frac{\sigma}{b}\ln \frac{e^{2j^*-1}}{(e-1)^{j^*-1}}\Bigg).
\end{align*}

By restricting the value of $\sigma$ under each scenario and combining the above results, we establish Theorem \ref{theorem:cr_rand}. Specifically, when $\sigma = \frac{1}{\tau^*}$, it can be verified that Theorem \ref{theorem:cr_rand} holds.
\end{proof}

\section{Proof for Algorithm 2}

\subsection{Proof of Lemma \ref{lemma:feasibility_learn}: Budget constraint} 
\label{app:budget_learn}
\begin{proof} We prove that the budget constraint is satisfied in expectation under each subroutine in Algorithm \ref{alg:pred}.

\textbf{Subroutine 4}
Let us suppose that $\tau = L + \beta e^{j^*}$ for some $j^* \in \mathbb{Z}^+$ and $\beta \in [b, be]$. Note that this implicitly implies that $\tau \geq L +b$, as we only consider the worst case where $\tau^*$ is large enough. Define $\delta = U-L$. Under the condition $U \leq b e$ or $\delta \leq b(e-1)$, we have $j^* = 0$. 


When $\delta \leq b$, Algorithm 2 would consistently use $p_{i} = \frac{b}{U}$, and the budget constraint is satisfied obviously. Now suppose $\delta > b$. When $\alpha \in[b, \beta]$, Algorithm 2 begins by setting $p_{i} = \frac{b}{\alpha +L}$ with a running length of $L + \alpha$ and then continues with $p_{i} = \frac{b}{U}$ for the second round with a running length of $L+\beta - L - \alpha$; when $\alpha \in [\beta, \delta]$, Algorithm 2 uses $p_{i} = \frac{b}{\alpha +L}$ with a running length of $L + \beta$ and stops; otherwise, the algorithm sets $p_{i} = \frac{b}{U}$ all the time. Therefore, the expected budget is




\begin{align*}
 \mathbb{E}&[\operatorname{Budget}] = \int_{b}^{\beta }\left[\frac{b}{L+\alpha}(L+\alpha) + \frac{b}{U}(L + \beta - L - \alpha)  \right]\frac{1}{\alpha} d\alpha + \int_{\beta }^{\delta} \frac{b}{L + \alpha}(L+\beta) \frac{1}{\alpha}d\alpha \\&\quad + \int_{\delta}^{b e}\frac{b}{U}(L+\beta) \frac{1}{\alpha}d\alpha\\
 &= b \ln \frac{\beta }{b} + \frac{b\beta}{U} \ln \frac{\beta}{b} - \frac{b}{U}(\beta -b)  + \frac{b (L + \beta)}{L}(\ln \frac{\delta}{\beta} - \ln \frac{L + \delta}{L + \beta}) + \frac{b (L + \beta)}{U} \ln \frac{b e}{\delta}\\
 &\leq b \ln \frac{U - L}{b} + \frac{b(U - L)}{U} \ln \frac{U-L}{b} - \frac{b}{U}(U-L-b)   + \frac{b U}{U} \ln \frac{b e}{\delta}\quad \text{increasing with} ~ \beta ~( \beta= U-L) \\
 &\leq b \ln (e-1) + \frac{b(b(e-1))}{L+b(e-1)} \ln (e-1) - \frac{b^2}{L+b(e-1)}(e-2)+ b \ln \frac{b e}{b(e-1)}\\
 &\quad \text{increasing with} ~ U ~( U= L + b(e-1))\\
 &\leq b \ln (e-1) + b\frac{e - 1}{e} \ln (e-1) - \frac{b}{e}(e-2))+ b \ln \frac{e}{e-1}\quad \text{decreasing with} ~ L ~( L=b)\\
 &\leq b + b(\frac{e-1}{e}\ln(e-1) - \frac{e-2}{e})\\
 &\approx b
\end{align*}

\textbf{Subroutine 5}

Similarly, we assume $\tau^* = L + \beta e^{j^*}$. Under the condition that $\delta \leq b(e+1)$, we have $j^* = 1$ or $j^* = 0$. As before, we only consider the worst case $j^*=1$. 

Let $\kappa = \frac{\delta}{e}$. Note that, when $\alpha \in [b, \kappa)$, Algorithm 2 first sets $p_{i} = \frac{b}{L + \alpha e}$ with a running length of $L+\alpha$, then transitions to $p_{k,i} = \frac{b}{U}$ with a running length of $L +\beta e - L - \alpha$. However, when $\alpha \in [\kappa, b e]$, Algorithm 2 keeps setting $p_{i} = \frac{b}{U}$ with a running length of $L+\beta e$. Therefore, the expected budget is
\begin{align*}
    \mathbb{E}&[\operatorname{Budget}] = \int_{b}^{\kappa} \left[\frac{b}{L+\alpha e}(L+\alpha) + \frac{b}{U}(L + \beta e - L - \alpha)\right]\frac{1}{\alpha}d\alpha + \int_{\kappa}^{b e} \frac{b}{U}(L +\beta e)\frac{1}{\alpha}d\alpha\\
    & = b\left(\ln \frac{\kappa}{b} + (\frac{1}{e}-1)\ln \frac{L+\kappa e}{L + b e}\right)+\frac{b}{U}\beta e\ln \frac{\kappa}{b} - \frac{b}{U}(\kappa - b) + \frac{b (\beta e + L)}{U} \ln \frac{b e}{\kappa}\\
    & \leq b\left(\ln \frac{\kappa}{b} + (\frac{1}{e}-1)\ln \frac{L+\kappa e}{L + b e}\right)+\frac{b}{U} (U-L)\ln \frac{\kappa}{b} - \frac{b}{U}(\kappa - b) + b \ln \frac{b e}{\kappa}\\
    &\quad \text{increasing with} ~ \beta ~( \beta= (U-L)/e)\\
    &\leq b + b (\frac{1}{e}-1)\ln \frac{L +b(e+1)}{L+b e} + \frac{b^2}{L+b(e+1)}(e+1)\ln \frac{e+1}{e} - \frac{b^2}{L+b(e+1)}(\frac{e+1}{e}-1) \\
    &\quad \text{increasing with} ~ U ~( U = L +b(e+1)) \\
    &\leq b + b (\frac{1-e}{e}\ln \frac{e+2}{e+1} + \frac{e+1}{e+2}\ln \frac{e+1}{e} - \frac{1}{e(e+2)})\quad \text{decreasing with} ~ L ~( L=b)\\
    &\approx b
\end{align*}

\textbf{Subroutine 6}

Under the assumption of $\tau^* = L + \beta e^{j^*}$, and given the condition $U > be^2$, it is possible for $j^*$ to be $ 0$ or to extend to infinity. We only focus on the worst-case scenario, i.e., $j^*\geq 1$. 

When $\alpha$ falls within $[b,\beta]$, Algorithm 2 stops with $p_i = b \left(1-\frac{L+\alpha-b}{L+\alpha(e-1)}\right)\frac{(1-1/e)^{j^*}}{\alpha e^{j^*+2}}$ with a running length of $L + \beta e^{j^*} - L - \alpha e^{j^*}$. Otherwise, Algorithm 2 stops with $p_i = b \left(1-\frac{L+\alpha-b}{L+\alpha(e-1)}\right)\frac{(1-1/e)^{j^*-1}}{\alpha e^{j^*+1}}$ with a running length of $L + \beta e^{j^*} - L - \alpha e^{j^*-1})$. Therefore, the expected budget is

\begin{align*}
    \mathbb{E}&[\operatorname{Budget}] = \int_{\beta}^{b e} \Bigg[\frac{b}{L + \alpha(e-1)}(L + \alpha) + b\left(1-\frac{L+\alpha-b}{L+\alpha(e-1)}\right)\sum_{j=2}^{j^*}\frac{(1-1/e)^{j-2}}{\alpha e^j}(\alpha e^{j-1} - \alpha e^{j-2}) \\
    &\quad + b \left(1-\frac{L+\alpha-b}{L+\alpha(e-1)}\right)\frac{(1-1/e)^{j^*-1}}{\alpha e^{j^*+1}}(L + \beta e^{j^*} - L - \alpha e^{j^*-1})\Bigg]\frac{1}{\alpha}d\alpha\\
    &\quad + \int_{b}^\beta \Bigg[\frac{b}{L + \alpha(e-1)}(L + \alpha)+ b\left(1-\frac{L+\alpha-b}{L+\alpha(e-1)}\right)\sum_{j=2}^{j^*+1}\frac{(1-1/e)^{j-2}}{\alpha e^j}(\alpha e^{j-1} - \alpha e^{j-2})\\
    &\quad + b \left(1-\frac{L+\alpha-b}{L+\alpha(e-1)}\right)\frac{(1-1/e)^{j^*}}{\alpha e^{j^*+2}}(L + \beta e^{j^*} - L - \alpha e^{j^*})\Bigg]\frac{1}{\alpha}d\alpha\\
    &\leq b \left(\ln\frac{b e}{\beta} + (\frac{1}{e-1} - 1)\ln \frac{L + b e(e-1)}{L + \beta (e-1)}\right) + b(1 - \frac{1}{e} - (1 - \frac{1}{e})^{j^*})\frac{e-2}{e-1} \ln \frac{L + b e(e-1)}{L + \beta (e-1)}\\
    &\quad + b(1 - \frac{1}{e} - (1 - \frac{1}{e})^{j^*})\frac{b}{L} \left(\ln \frac{b e}{\beta} - \frac{L + b e(e-1)}{L + \beta (e-1)}\right)\\
    &\quad + b \beta e^{j^*} \frac{(e-1)^{j^*}}{e^{2j^*}} \frac{1}{L}\left(\ln \frac{b e}{\beta} - \ln \frac{L + be(e-1)}{L + \beta(e-1)}\right) - b \frac{(e-1)^{j^*-1}}{e^{j^* + 1}}\ln \frac{L + b e(e-1)}{L+\beta (e-1)}\\
    &\quad + b \left(\ln\frac{\beta}{b} + (\frac{1}{e-1}-1)\ln \frac{L + \beta (e-1)}{L + b(e-1)}\right) + b (1-\frac{1}{e}-(1-\frac{1}{e})^{j^*+1})\frac{e-2}{e-1} \ln \frac{L +\beta(e-1)}{L+b(e-1)}\\
    &\quad + b(1 - \frac{1}{e} - (1 - \frac{1}{e})^{j^*})\frac{b}{L} \left(\ln \frac{\beta}{b} - \frac{L + \beta (e-1)}{L + b (e-1)}\right)\\
    &\quad + b \beta e^{j^*}\frac{(e-1)^{j^*+1}}{e^{2j^*+2}} \frac{1}{L}\left(\ln \frac{\beta}{b} - \ln \frac{L + \beta (e-1)}{L + b(e-1)}\right) - b\frac{(e-1)^{j^*}}{e^{j^* + 2}}\ln \frac{L +\beta (e-1)}{L + b(e-1)}\\
    &\leq b \left(1 + (\frac{1}{e-1}-1)\ln \frac{L + be (e-1)}{L + b(e-1)}\right) + b (1-\frac{1}{e}-(1-\frac{1}{e})^{j^*+1})\frac{e-2}{e-1} \ln \frac{L +b e(e-1)}{L+b(e-1)}\\
    &\quad + b(1 - \frac{1}{e} - (1 - \frac{1}{e})^{j^*})\frac{b}{L} \left(1 - \frac{L + b e (e-1)}{L + b (e-1)}\right)\\
    &\quad + b^2 \frac{(e-1)^{j^*+1}}{e^{j^*+1}} \frac{1}{L}\left(1 - \ln \frac{L + b e (e-1)}{L + b(e-1)}\right) - b\frac{(e-1)^{j^*}}{e^{j^* + 2}}\ln \frac{L +b e (e-1)}{L + b(e-1)}\\
    &\quad \text{increasing with} ~ \beta ~(\beta = be)\\
    &\leq b \left(1 + (\frac{1}{e-1}-1)\ln \frac{L + be (e-1)}{L + b(e-1)}\right) + b\frac{e-2}{e}\ln \frac{L + be (e-1)}{L + b(e-1)} \\&\quad + b(1-\frac{1}{e})\frac{b}{L}\left(1 - \ln\frac{L + b e (e-1)}{L + b (e-1)}\right)\\
    &\approx b
\end{align*}

Combining the above results with the proof of Subroutine 2, presented in Appendix \ref{app:budget_rand}, establishes Lemma \ref{lemma:feasibility_learn}.
\end{proof}

\subsection{Proof of Theorem \ref{theorem:cr_learn}: Consistency and Robustness}
\label{app:cr_learn}
\begin{proof}   We begin with the proof of consistency and then proceed to the analysis of robustness.

\textbf{Consistency Analysis}
It is straightforward to show that our algorithm is $1$- consistent. When the width of the predictive interval is zero, meaning that $L = U = \tau^*$, we have
\begin{align*}
    \mathbb{E}[\operatorname{SOL}] = \frac{b}{U} \tau^* = b.
\end{align*}

\textbf{Robustness Analysis}
Below we show the robustness of our algorithm under each subroutine.

\textbf{Subroutine 4} 

For cases where $\delta  = U-L \leq b$,  the algorithm proceeds with $p_{k,i} = \frac{b}{U}$. Hence, we have
\begin{align*}
    \mathbb{E}[\operatorname{SOL}] = \frac{b}{U}\beta \geq \frac{b}{L + b} L \geq \frac{b}{2}.
\end{align*}

Next, we consider the case where $b(e-1) \geq \delta > b$, further divided into $\tau^* < L +b$ and $\tau^* \geq L + b$. 

Suppose $\tau^* < L +b$. When $\alpha$ falls within $[b, \delta]$, Algorithm 2 assigns $p_{i} = \frac{b}{L + \alpha}$ with a running length of $\tau^*$. Otherwise, Algorithm 2 sets $p_{i} = \frac{b}{U}$ with a running length of $\tau^*$. Therefore, the expected solution is
\begin{align*}
    \mathbb{E}[\operatorname{SOL}] &= \int_{b}^{\delta}\left[\frac{b}{L+\alpha}\tau^*\right]\frac{1}{\alpha}d\alpha + \int_{\delta}^{be} \frac{b}{U}\tau^*\frac{1}{\alpha}d\alpha\\
    &= \frac{b\tau^*}{L}\left(\ln \frac{\delta}{b} - \ln \frac{L+\delta}{L+b}\right) + \frac{b \tau^*}{U}\ln \frac{be}{\delta}\\
    &\geq b\left(\ln \frac{\delta}{b} - \ln \frac{L+\delta}{L+b}\right) + \frac{b L}{U}\ln \frac{be}{\delta}\quad\text{increasing with} ~ \tau^* ~( \tau^* =L )\\
    &\geq b\left(\ln (e-1) - \ln \frac{L+b(e-1)}{L+b}\right) + \frac{bL}{L + b(e-1)}\ln \frac{e}{e-1} \\&\quad \text{decreasing with} ~ U ~( U=L + b(e-1))\\
    &\geq b\left(\ln (e-1) - \ln \frac{e}{2}\right) + b\frac{1}{e}\ln \frac{e}{e-1} \quad \text{increasing with} ~ L ~( L = b)\\
    &= b\left(\ln \frac{2(e-1)}{e}+\frac{1}{e}\ln\frac{e}{e-1}\right)
\end{align*}

For cases where $\tau^* \geq L + b$, let us suppose $\tau^* = L + \beta e^{j^*}$ where $\beta \in [b, b e]$. Under the condition $U \leq b e$ or $\delta \leq b(e-1)$, we have $j^* = 0$. Further, since $L +\beta \leq U$, we have $\beta \leq U-L$. When $\alpha$ falls within $[b,\beta]$, Algorithm 2 starts with $p_i = \frac{b}{L+\alpha}$ with a running length of $L + \alpha$, then transitions to $p_i = \frac{b}{U}$ with a running length of $L+\beta - L - \alpha$; when $\alpha$ falls within $[\beta, \delta]$, Algorithm 2 assigns $p_i = \frac{b}{L+\alpha}$ with a running length of $L+\beta$; otherwise, Algorithm 2 assigns $p_i = \frac{b}{U}$ with a running length of $L+\beta$. 
It follows that
\begin{align*}
    &\mathbb{E}[\operatorname{SOL}] = \int_{b}^{\beta }\left[\frac{b}{L+\alpha}(L+\alpha) + \frac{b}{U}(L + \beta - L - \alpha) - \sigma   \ln \frac{U}{L+\alpha} \right]\frac{1}{\alpha} d\alpha\\
    &\quad + \int_{\beta }^{\delta} \frac{b}{L + \alpha}(L +\beta) \frac{1}{\alpha}d\alpha + \int_{\delta}^{b e}\frac{b}{U}(L +\beta)\frac{1}{\alpha}d\alpha\\
    &= b\ln \frac{\beta }{b} + \frac{b \beta}{U}\ln \frac{\beta}{b} -\frac{b}{U}(\beta - b) - \sigma \ln \frac{e}{2} \ln \frac{\beta}{b}  + \frac{b (L+\beta)}{L}(\ln \frac{\delta}{\beta}  - \ln \frac{L + \delta}{L + \beta}) + \frac{b (L + \beta) }{U} \ln \frac{b e}{\delta}\\
    &\geq \frac{b (L+b)}{L}\left(\ln \frac{U-L}{b}  - \ln \frac{U}{L + b}\right) + \frac{b (L + b) }{U} \ln \frac{b e}{U-L}\quad \text{increasing with} ~ \beta ~( \beta = b)\\
    &\geq \frac{b (L+b)}{L} \left(\ln (e-1) - \ln \frac{L + b(e-1)}{L + b}\right) + \frac{b(L+b)}{L+b(e-1)}\ln \frac{e}{e-1}\\
    &\quad \text{decreasing with} ~ U ~( U=L + b(e-1))\\
    &\geq 2b \ln \frac{2(e-1)}{e} + b\frac{2}{e}\ln \frac{e}{e-1}\quad \text{increasing with} ~ L ~( L= b)\\
    &= b (2\ln \frac{2(e-1)}{e} + \frac{2}{e}\ln \frac{e}{e-1})
\end{align*}



\textbf{Subroutine 5} 

For the situation where $\delta = U-L \leq b e$, the algorithm's procedure involves setting $p_{i}=\frac{b}{U}$ for all time points. Hence, we have
\begin{align*}
    \mathbb{E}[\operatorname{SOL}] = \frac{b}{U}\tau^* \geq \frac{b}{L + be} L \geq \frac{b}{e+1}.
\end{align*}

Next, we consider the case where $b (e+1) \geq \delta > be$, further divided into $\tau^* < L+b$ and $\tau^* \geq L + b$. 

First, suppose $\tau^* < L+b$. When $\alpha$ falls in the range $[b,\kappa]$, Algorithm 2 assigns $p_i = \frac{b}{L+\alpha e}$ with a running length $\tau^*$. Otherwise, Algorithm 2 assigns $p_i = \frac{b}{U}$ with a running length $\tau^*$. Therefore, the expected solution is
\begin{align*}
    \mathbb{E}[\operatorname{SOL}] &= \int_{b}^{\kappa} \left[\frac{b}{L + \alpha e}\tau^*\right] \frac{1}{\alpha} d\alpha + \int_{\kappa}^{b e} \frac{b}{U}\tau^* \frac{1}{\alpha} d\alpha\\
    &= \frac{b \tau^*}{L}\left(\ln\frac{\kappa}{b}-\ln \frac{L + \kappa e}{L + be}\right)  + \frac{b \tau^*}{U}\ln \frac{be}{\kappa}\\
    &\geq b\left(\ln\frac{\kappa}{b}-\ln \frac{L + \kappa e}{L + be}\right)  + \frac{b L}{U}\ln \frac{be}{\kappa}\quad \text{increasing with} ~ \tau^* ~( \tau^*=L) \\
     &\geq  b \left(\ln\frac{e+1}{e}-\ln \frac{L + b(e+1)}{L + be}\right) + \frac{b L}{L+b(e+1)}\ln \frac{e^2}{e+1}\\&\quad \text{decreasing with} ~ U ~( U=L + b(e+1))\\
     &\geq b\left(\ln\frac{e+1}{e}-\ln \frac{e^2}{e^2-1}\right) + b\frac{e^2-e-1}{e^2}\ln \frac{e^2}{e+1}
     \\&\quad \text{increasing with} ~ L ~( L=b(e^2-e-1))\\
     &= b \left(\ln\frac{e+1}{e}-\ln \frac{e^2}{e^2-1} + \frac{e^2-e-1}{e^2}\ln \frac{e^2}{e+1}\right)
\end{align*}

For situations where $\tau^* \geq L +b$, suppose $\tau^* = L + \beta e^{j^*}$ where $\beta \in [b, b e]$. Below we separately consider two cases: 1) $j^* \geq 1$ or $\beta \geq \kappa$, and 2) $j^* = 0$ and $\beta < \kappa$.

Suppose case 1) where $j^* \geq 1$ or $\beta \geq \kappa$. When $\alpha$ falls within $[b,\kappa]$, Algorithm 2 first assigns $p_{i} = \frac{b}{L+\alpha e}$ for a time length of $L + \alpha$, and then proceeds with $p_{k,i} = \frac{b}{U}$ with a running length $L + \beta e^{j^*} - L - \alpha$. Otherwise, Algorithm 2 assigns $p_i = \frac{b}{U}$ with a running length of $L + \beta e^{j^*}$.
Therefore, the expected solution is
\begin{align*}
     &\mathbb{E}[\operatorname{SOL}] = \int_{b}^{\kappa}\left[\frac{b}{L+\alpha e}(L+\alpha) + \frac{b}{U}(\beta e^{j^*} + L - L - \alpha) - \sigma \ln \frac{U}{L+\alpha e}\right]\frac{1}{\alpha}d\alpha 
     \\&\quad+ \int_{\kappa}^{b e}\frac{b}{U}(\beta e^{j^*} + L)\frac{1}{\alpha}d\alpha\\
     & \geq b\left(\ln \frac{\kappa}{b} + (\frac{1}{e}-1)\ln \frac{L+\kappa e}{L + b e}\right)+\frac{b}{U}\beta e^{j^*}\ln \frac{\kappa}{b} - \frac{b}{U}(\kappa - b)- \sigma \ln \frac{e + 2}{e+1} \ln \frac{\kappa}{b} \\&\quad+ \frac{b (\beta e^{j^*} + L)}{U} \ln \frac{b e}{\kappa}\\
     & \geq b\left(\ln \frac{\kappa}{b} + (\frac{1}{e}-1)\ln \frac{L+\kappa e}{L + b e}\right)+\frac{b}{U}\kappa\ln \frac{\kappa}{b} - \frac{b}{U}(\kappa - b)- \sigma\ln \frac{e + 2}{e+1} \ln \frac{\kappa}{b} \\&\quad + \frac{b (\kappa + L)}{U} \ln \frac{b e}{\kappa} \quad
     \text{increasing with} ~ \beta, j^* ~( \beta = \kappa, j^*=0) \\
     &\geq b \left(\ln \frac{e+1}{e} - \frac{e-1}{e}\ln \frac{L +b(e+1)}{L+be}\right) + \frac{b^2(e+1)}{Le+be(e+1)} \ln \frac{e+1}{e}- \frac{b}{L+b(e+1)}(\frac{b(e+1)}{e}-b)\\
     &\quad  - \sigma \ln \frac{e+2}{e+1}\ln \frac{e+1}{e} + \frac{b(L + \frac{b(e+1)}{e} }{L+b(e+1)}\ln \frac{e^2}{e+1}\quad \text{decreasing with} ~ U ~( U=L + b(e+1)) \\
     &\geq b(\ln \frac{e+1}{e} - \frac{e-1}{e}\ln \frac{e^2}{e^2-1}) + \frac{b(e+1)}{e^3}\ln \frac{e+1}{e} - b\frac{1}{e^2}\frac{1}{e}- \sigma \ln \frac{e+2}{e+1}\ln \frac{e+1}{e}\\
     &\quad  + b\frac{e^2-e-1 +\frac{(e+1)}{e}}{e^2}\ln\frac{e^2}{e+1}\quad \text{increasing with} ~ L ~( L= b(e^2-e-1))\\
     &= b\left(\ln \frac{e+1}{e} - \frac{1}{e^3} - \frac{e-1}{e}\ln \frac{e^2}{e^2-1} +\frac{e^2-e-1}{e^2}\ln \frac{e^2}{e+1}  \right) + b\frac{1+\frac{1}{e}}{e^2}
     \\&\quad - \sigma\ln \frac{e+2}{e+1}\ln \frac{e+1}{e}\\
     &= b \left((1+\frac{1}{e^2})\ln(e+1) - 1 - \frac{1}{e^2} + \frac{e-1}{e}\ln(e-1)\right)- \sigma \ln \frac{e+2}{e+1}\ln \frac{e+1}{e}
\end{align*}

Next, consider case 2) where $j^* = 0$ and $\beta < \kappa$. When $\alpha$ falls within $[b,\beta]$, Algorithm 2 starts with $p_i = \frac{b}{L+\alpha e}$ with a running length of $L + \alpha$, then transits to $p_i = \frac{b}{U}$ with a running length of $L+\beta - L-\alpha$; when $\alpha \in [\beta, \kappa]$, Algorithm 2 sets $p_i = \frac{b}{L+\alpha e}$ with a running length of $L + \beta$; otherwise, Algorithm 2 sets $p_i = \frac{b}{U}$ with a running length of  $L + \beta$.
Therefore, we have 
\begin{align*}
    \mathbb{E}[\operatorname{SOL}] &= \int_{b}^{\beta}\left[\frac{b}{L+\alpha e}(L+\alpha) + \frac{b}{U}(\beta + L - L - \alpha) - \sigma  \ln \frac{U}{L+\alpha e}\right]\frac{1}{\alpha}d\alpha\\
    &\quad + \int_{\beta }^{\kappa}\frac{b}{L+\alpha e}(L + \beta)\frac{1}{\alpha}d\alpha + \int_{\kappa}^{b e}\frac{b}{U}(L + \beta) \frac{1}{\alpha}d\alpha\\
    &=b\left(\ln \frac{\beta }{b} + (\frac{1}{e}-1)\ln \frac{L+\beta e}{L + be}\right) + \frac{b}{U}\beta \ln \frac{\beta }{b}  - \frac{b}{U}(\beta  - b) - \sigma \ln \frac{e^2}{e^2- 1} \ln \frac{\beta}{b} \\
    &\quad + \frac{b (L + \beta) }{L}\left(\ln\frac{\kappa}{\beta }-\ln \frac{L + \kappa e}{L + \beta e}\right) + \frac{b (L +\beta)}{U}\ln \frac{be}{\kappa}\\
    &\geq \frac{b (L + \beta) }{L}\left(\ln\frac{\kappa}{\beta }-\ln \frac{L + \kappa e}{L + \beta e}\right) + \frac{b (L +\beta)}{U}\ln \frac{be}{\kappa}\quad \text{increasing with} ~ \beta ~( \beta=b)\\
    &\geq \frac{b (L +b)}{L}\left(\ln \frac{b(e+1)}{b e} - \ln \frac{L+b(e+1)}{L+be}\right) + \frac{b (L +b)}{L+b(e+1)}\ln\frac{e^2}{e+1}\\&\quad \text{decreasing with} ~ U ~( U=L + b(e+1)) \\
    &\geq b\frac{e^2-e}{e^2-e-1}\left(\ln \frac{e+1}{e} - \ln \frac{e^2}{e^2-1}\right) + \frac{e^2-e}{e^2}\ln\frac{e^2}{e+1}\\
    &\quad \text{increasing with} ~ L ~( L= b(e^2-e-1))\\
    &\geq b\left(\frac{e^2-e}{e^2-e-1}\ln \frac{(e+1)^2(e-1)}{e^3} +\frac{e-1}{e} \ln \frac{e^2}{e+1}\right).
\end{align*}

\textbf{Subroutine 6}

In this scenario, our algorithm initiates with $p_{k,i} = \frac{b}{L + \alpha(e-1)}$, subsequently updating $\tilde{\tau}$ and $b$ after each iteration. We analyze two cases: one where $\tau^* < L + b$, and the other where $\tau^* \geq L + b$.

For the first situation where $\tau^* < L + b$, Algorithm 2 consistently sets $p_i = \frac{b}{L+\alpha(e-1)}$. Therefore, we have
\begin{align*}
     \mathbb{E}[\operatorname{SOL}] &= \int_{b}^{b e} \frac{b}{L+\alpha(e-1)} \tau^* \frac{1}{\alpha}d\alpha \\
     &=\frac{b \tau^*}{L}\left(\ln\frac{b e}{b}-\ln \frac{L + b e(e-1)}{L + b(e-1)}\right) \\
     &\geq \tau^* \left(1-\ln \frac{e^2-e+1}{e}\right) \quad \text{increasing with} ~ L ~( L=b)\\
     &\geq b\left(2-\ln (e^2-e+1)\right)
\end{align*}

Next, we consider the case where $\tau^* \geq L + b$. Suppose that $\tau^* = L + \beta e^{j^*}$ where $\beta \in [b, b e]$.

When $j^* \geq 1$, the objective function when $\alpha \geq \beta$ is
\begin{align*}
    \operatorname{SOL}_{1} &= \frac{b}{L+\alpha(e-1)} (L +\alpha) + b \left(1 - \frac{L+\alpha - b}{L + \alpha (e-1)}\right)\sum_{j=2}^{j^*} \frac{(1-1/e)^{j-2}}{\alpha e^j}(\alpha e^{j-1} - \alpha e^{j-2})\\
    &\quad + b\left(1 - \frac{L+\alpha-b}{L + \alpha (e-1)}\right) \frac{(1-1/e)^{j^* - 1}}{\alpha e^{j^*}}(L + \beta e^{j^*}- L - \alpha e^{j^*- 1}) \\
    &\quad- \sigma \ln\frac{\alpha e^{2j^*+1}}{(\alpha(e-2) + b)(e-1)^{j^*-1}}
\end{align*}
The objective function when $\alpha < \beta$ is
\begin{align*}
    \operatorname{SOL}_{2} &= \frac{b}{L+\alpha(e-1)} (L +\alpha) + b \left(1 - \frac{L+\alpha - b}{L + \alpha (e-1)}\right)\sum_{j=2}^{j^*+1} \frac{(1-1/e)^{j-2}}{\alpha e^j}(\alpha e^{j-1} - \alpha e^{j-2})\\
    &\quad + b\left(1 - \frac{L+\alpha-b}{L + \alpha (e-1)}\right) \frac{(1-1/e)^{j^*}}{\alpha e^{j^*+2}}(L + \beta e^{j^*}- L - \alpha e^{j^*}) \\&\quad - \sigma \ln\frac{\alpha e^{2j^*+2}}{(\alpha(e-2)+b) (e-1)^{j^*}}.
\end{align*}

The expected solution is
\begin{align*}
    &\mathbb{E}[\operatorname{SOL}] = \int_{\beta}^{b e} \operatorname{SOL}_1 \frac{1}{\alpha} d\alpha + \int_{b}^\beta \operatorname{SOL}_2 \frac{1}{\alpha} d\alpha \\
    &\geq b \left(\ln\frac{b e}{\beta} + (\frac{1}{e-1} - 1)\ln \frac{L + b e(e-1)}{L + \beta (e-1)}\right) + b(1 - \frac{1}{e} - (1 - \frac{1}{e})^{j^*})\frac{e-2}{e-1} \ln \frac{L + b e(e-1)}{L + \beta (e-1)}\\
    &\quad + b^2(1 - \frac{1}{e} - (1 - \frac{1}{e})^{j^*})\frac{1}{L}\left(\ln \frac{be}{\beta} -\ln \frac{L + b e(e-1)}{L + \beta (e-1)}\right)\\
    &\quad + b \beta e^{j^*} \frac{(e-1)^{j^*-1}}{e^{2j^*}} \frac{e-2}{L}\left(\ln \frac{b e}{\beta} - \ln \frac{L + be(e-1)}{L + \beta(e-1)}\right) - b \frac{(e-1)^{j^*-1}}{e^{j^* + 1}}\frac{e-2}{e-1}\ln \frac{L + b e(e-1)}{L+\beta (e-1)}\\
    &\quad- \sigma \ln\frac{e^{2j^*+1}}{(e-1)^{j^*+1}}\ln \frac{be}{\beta} \\
    &\quad + b \left(\ln\frac{\beta}{b} + (\frac{1}{e-1}-1)\ln \frac{L + \beta (e-1)}{L + b(e-1)}\right) + b (1-\frac{1}{e}-(1-\frac{1}{e})^{j^*+1})\frac{e-2}{e-1} \ln \frac{L +\beta(e-1)}{L+b(e-1)}\\
    &\quad + b^2 (1-\frac{1}{e}-(1-\frac{1}{e})^{j^*+1})\frac{1}{L} \left(\ln\frac{\beta}{b} -\ln \frac{L +\beta(e-1)}{L+b(e-1)}\right)\\
    &\quad + b \beta e^{j^*}\frac{(e-1)^{j^*}}{e^{2j^*+2}} \frac{e-2}{L}\left(\ln \frac{\beta}{b} - \ln \frac{L + \beta (e-1)}{L + b(e-1)}\right) - b\frac{(e-1)^{j^*}}{e^{j^* + 2}}\frac{e-2}{e-1}\ln \frac{L +\beta (e-1)}{L + b(e-1)}\\
    &\quad - \sigma \ln \frac{e^{2j^*+3}}{(e-1)^{j^*+2}}\ln \frac{\beta}{b}\\
    &\geq b \left(1 + (\frac{1}{e-1} - 1)\ln \frac{L + b e(e-1)}{L + b (e-1)}\right) + b(1 - \frac{1}{e} - (1 - \frac{1}{e})^{j^*})\frac{e-2}{e-1} \ln \frac{L + b e(e-1)}{L + b (e-1)}\\
    &\quad+ b^2(1 - \frac{1}{e} - (1 - \frac{1}{e})^{j^*})\frac{1}{L}\left(1 -\ln \frac{L + b e(e-1)}{L + b (e-1)}\right)\\
    &\quad + b^2 e^{j^*} \frac{(e-1)^{j^*-1}}{e^{2j^*}} \frac{e-2}{L}\left(1 - \ln \frac{L + be(e-1)}{L + b(e-1)}\right) - b \frac{(e-1)^{j^*-1}}{e^{j^* + 1}}\frac{e-2}{e-1}\ln \frac{L + b e(e-1)}{L+b (e-1)}\\
    &\quad- \sigma \ln\frac{e^{2j^*+1}}{(e-1)^{j^*+1}} \quad \text{increasing with} ~ \beta ~(\beta=b)\\
    &\geq b \left(1 + (\frac{1}{e-1} - 1)\ln \frac{b (e^2 -e+1)}{b + b (e-1)}\right) + b(1 - \frac{1}{e} - (1 - \frac{1}{e})^{j^*})\frac{e-2}{e-1}  \ln \frac{b (e^2-e+1)}{b + b (e-1)}\\
    &\quad+ b(1 - \frac{1}{e} - (1 - \frac{1}{e})^{j^*})\left(1 -\ln \frac{e^2-e+1}{e}\right)\\
    &\quad + b e^{j^*} \frac{(e-1)^{j^*-1}}{e^{2j^*}}(e-2) \left(1 - \ln \frac{b(e^2-e+1)}{b + b(e-1)}\right) - b \frac{(e-1)^{j^*-1}}{e^{j^* + 1}}\frac{e-2}{e-1}\ln \frac{e^2-e+1}{e}\\
    &\quad - \sigma \ln\frac{e^{2j^*+1}}{(e-1)^{j^*+1}} \quad \text{increasing with} ~ L ~( L=b)\\
    &= b \left(1 + (\frac{1}{e-1} - 1)\ln \frac{e^2 -e+1}{e}\right) + b(1 - \frac{1}{e} - (1 - \frac{1}{e})^{j^*})\frac{e-2}{e-1} \ln \frac{e^2-e+1}{e}\\
    &\quad+ b(1 - \frac{1}{e} - (1 - \frac{1}{e})^{j^*})\left(1 -\ln \frac{e^2-e+1}{e}\right)\\
    &\quad + b \frac{(e-1)^{j^*-1}}{e^{j^*}} (e-2)\left(1 - \ln \frac{e^2-e+1}{e}\right) - b \frac{(e-1)^{j^*-1}}{e^{j^* + 1}}\frac{e-2}{e-1}\ln \frac{e^2-e+1}{e}\\& \quad - \sigma \ln\frac{e^{2j^*+1}}{(e-1)^{j^*+1}}. 
\end{align*}

Next, consider the case where $j^* =0$. When $\alpha \in [b,\beta]$, Algorithm 2 starts with $p_i = \frac{b}{L+\alpha (e-1)}$ with a running length of $L+\alpha$, then transitions to $p_i = b\left(1-\frac{L+\alpha-b}{L+\alpha(e-1)}\right)\frac{1}{\alpha e^2}$ with a running length of $L +\beta -L-\alpha$. Otherwise, Algorithm 2 consistently sets $p_i = \frac{b}{L+\alpha (e-1)}$. Therefore, we have
\begin{align*}
    &\mathbb{E}[\operatorname{SOL}] = \int_{b}^{\beta }\Bigg[\frac{b}{L + \alpha(e-1)} (L + \alpha) +b \left(1 - \frac{L+\alpha - b}{L + \alpha(e-1)}\right) \frac{1}{\alpha e^2} (L + \beta - L - \alpha)\\&\quad - \sigma \ln \frac{\alpha e^2}{\alpha (e-2) + b} \Bigg] \frac{1}{\alpha} d\alpha
     + \int_{\beta}^{be} \frac{b}{L+\alpha(e-1)}(L +\beta) \frac{1}{\alpha}d\alpha\\
    & \geq b\left(\ln \frac{\beta }{b} - \frac{e-2}{e-1}\ln \frac{L+\beta(e-1)}{L + b(e-1)}\right) - \frac{b \beta}{e^2}\frac{e-2}{L}\left(\ln \frac{\beta}{b}-\ln \frac{L + \beta(e-1)}{L+b(e-1)}\right) \\
    &\quad -\frac{b}{e^2}\frac{e-2}{e-1}\ln \frac{L + \beta(e-1)}{L + b(e-1)} - \sigma \ln \frac{e^3}{(e-1)^2}\ln \frac{\beta}{b} + \frac{b (L +\beta)}{L}\left(\ln \frac{b e}{\beta } - \ln \frac{L + b e(e-1)}{L+\beta(e-1)}\right)\\
    &\geq \frac{b (L +b)}{L}\left(1- \ln \frac{L + b e(e-1)}{L+b(e-1)}\right)\quad \text{increasing with} ~ \beta ~(\beta=b)\\
    &\geq  2b \left(1- \ln \frac{b(e^2 - e+1)}{b+b(e-1)}\right) \quad \text{increasing with} ~ L ~( L=b)\\
    &= 2b \left(2-\ln (e^2-e+1)\right).
\end{align*}

\paragraph{Tuning parameter selection}
For Scenario 1) where $U \leq be$, the competitive ratio is 
\begin{align*}
\ln \frac{2(e-1)}{e}+\frac{1}{e}\ln \frac{e}{e-1}.    
\end{align*}

For Scenario 2) where $be < U \leq be^2$, the competitive ratio  is
\begin{align*}
    \min \left(\frac{1}{e}, \frac{2}{e} - \frac{\sigma}{b} \right).
\end{align*}

For Scenario 3) where $U > be^2$, the competitive ratio is
\begin{align*}
    \min \Bigg(2-\ln(e^2-e+1),  &1 + \left(\frac{1}{e-1} - 1\right)\ln \frac{e^2 -e+1}{e} + (1 - \frac{1}{e} - (1 - \frac{1}{e})^{j^*})\frac{e-2}{e-1} \ln \frac{e^2-e+1}{e}\\
    &\quad+ (1 - \frac{1}{e} - (1 - \frac{1}{e})^{j^*})\left(1 -\ln \frac{e^2-e+1}{e}\right)+ \frac{(e-1)^{j^*-1}}{e^{j^*}} (e-2)\left(1 - \ln \frac{e^2-e+1}{e}\right)\\
    &\quad  - \frac{(e-1)^{j^*-1}}{e^{j^* + 1}}\frac{e-2}{e-1}\ln \frac{e^2-e+1}{e} - \frac{\sigma}{b} \ln\frac{e^{2j^*+1}}{(e-1)^{j^*+1}}\Bigg).
\end{align*}

By restricting the value of $\sigma$ under each scenario and combining the above results, we establish Theorem \ref{theorem:cr_learn}. Specifically, when $\sigma = \frac{1}{\tau^*}$, it can be verified that Theorem \ref{theorem:cr_learn} holds.
\end{proof}

\section{Additional Synthetic Experiments}

\subsection{Performance under Small $\tau^*$}
\label{app:results_smalltau}
\begin{figure*}[ht!]
    \centering
    \begin{subfigure}[b]{0.31\linewidth}
        \centering
        \includegraphics[width=\linewidth,keepaspectratio]{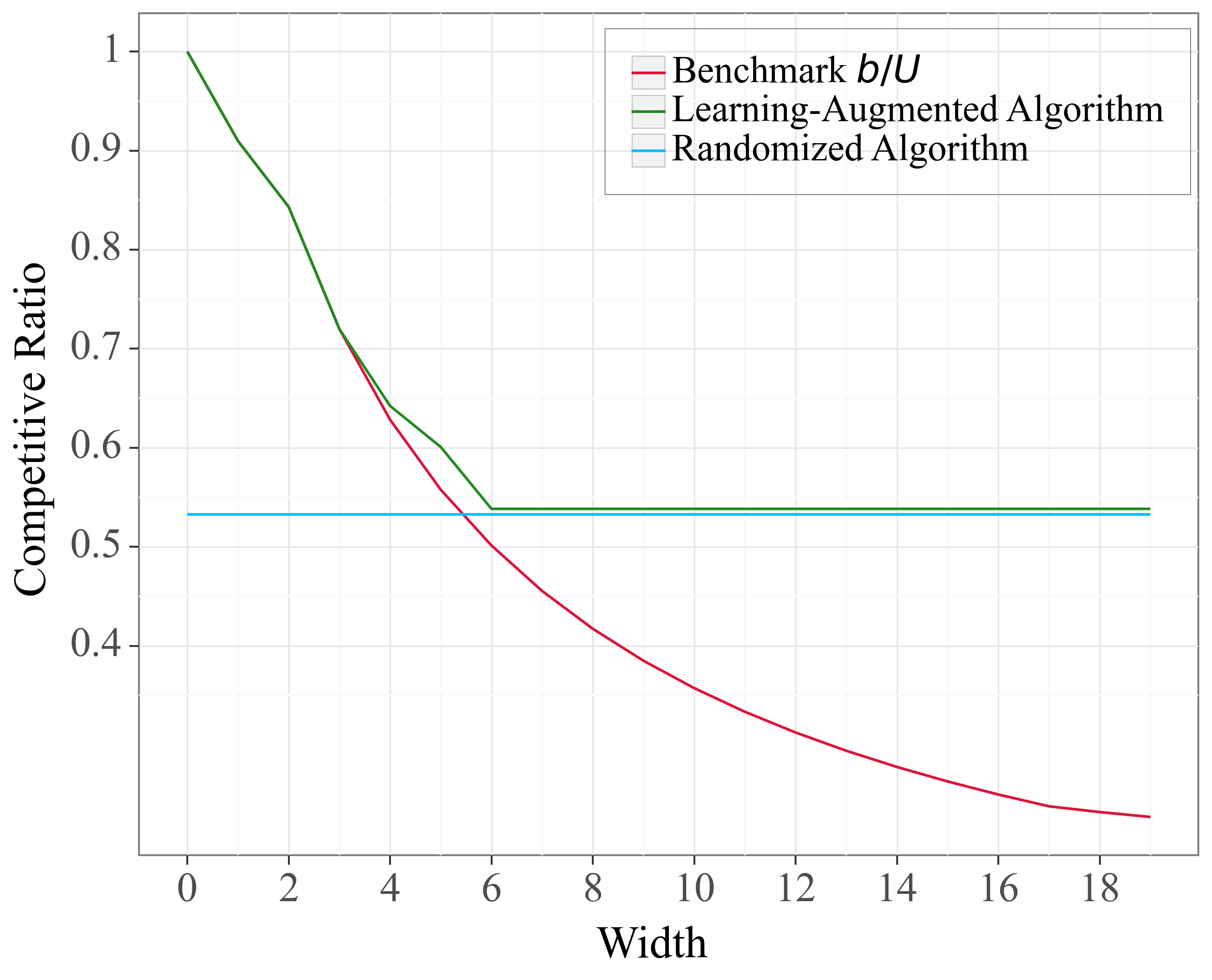}
        \vspace{-17pt}
        \caption{Scenario 1: $T=22$, $\tau^* = \text{Int}[0.2(T+b)]$}
        \label{fig:set22_a}
    \end{subfigure}
    \hfill
    \begin{subfigure}[b]{0.31\linewidth}
        \centering
        \includegraphics[width=\linewidth,keepaspectratio]{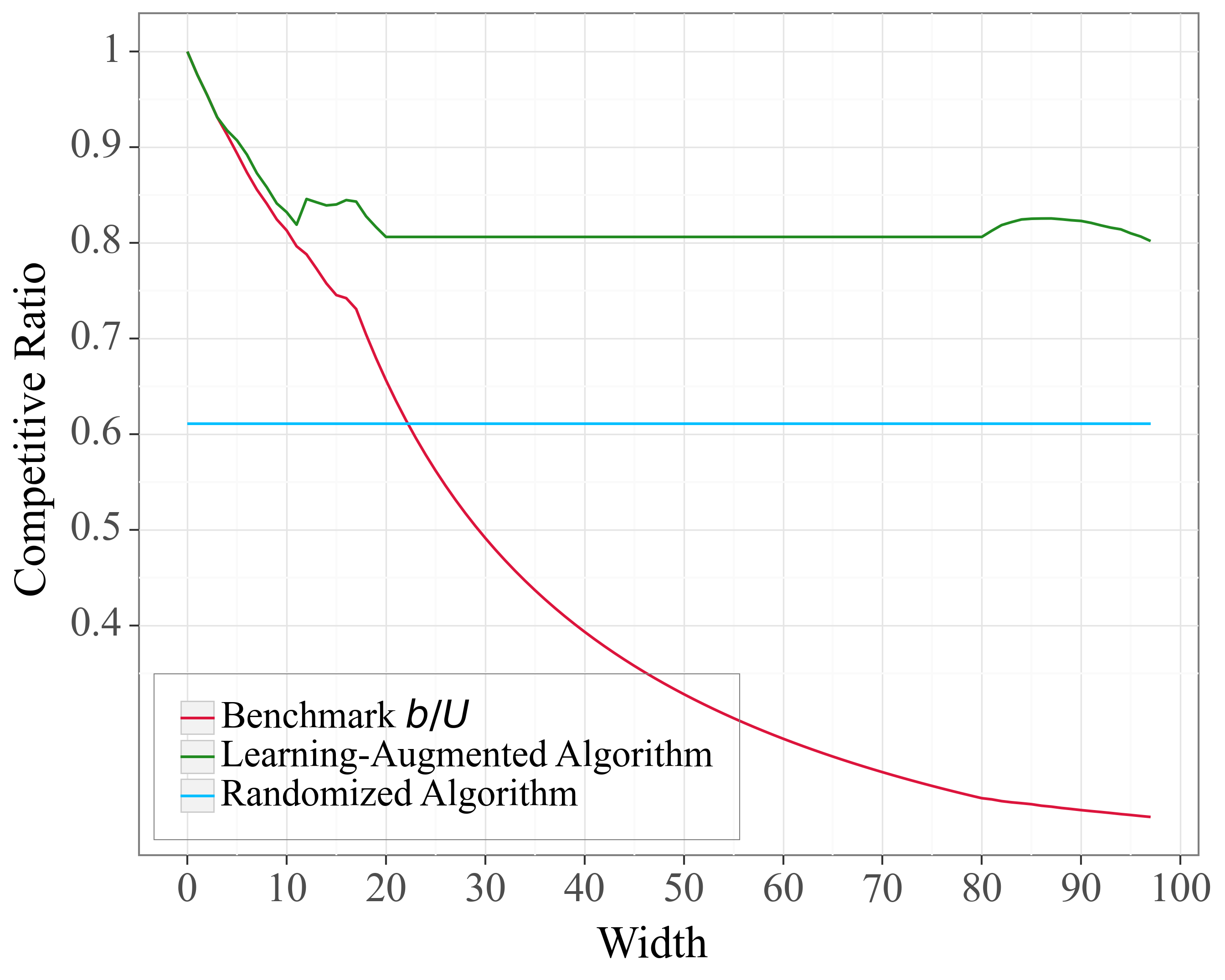}
        \vspace{-17pt}
        \caption{Scenario 2: $T=100$, $\tau^* = \text{Int}[0.2(T+b)]$}
        \label{fig:set22_b}
    \end{subfigure}
    \hfill
    \begin{subfigure}[b]{0.31\linewidth}
        \centering
        \includegraphics[width=\linewidth,keepaspectratio]{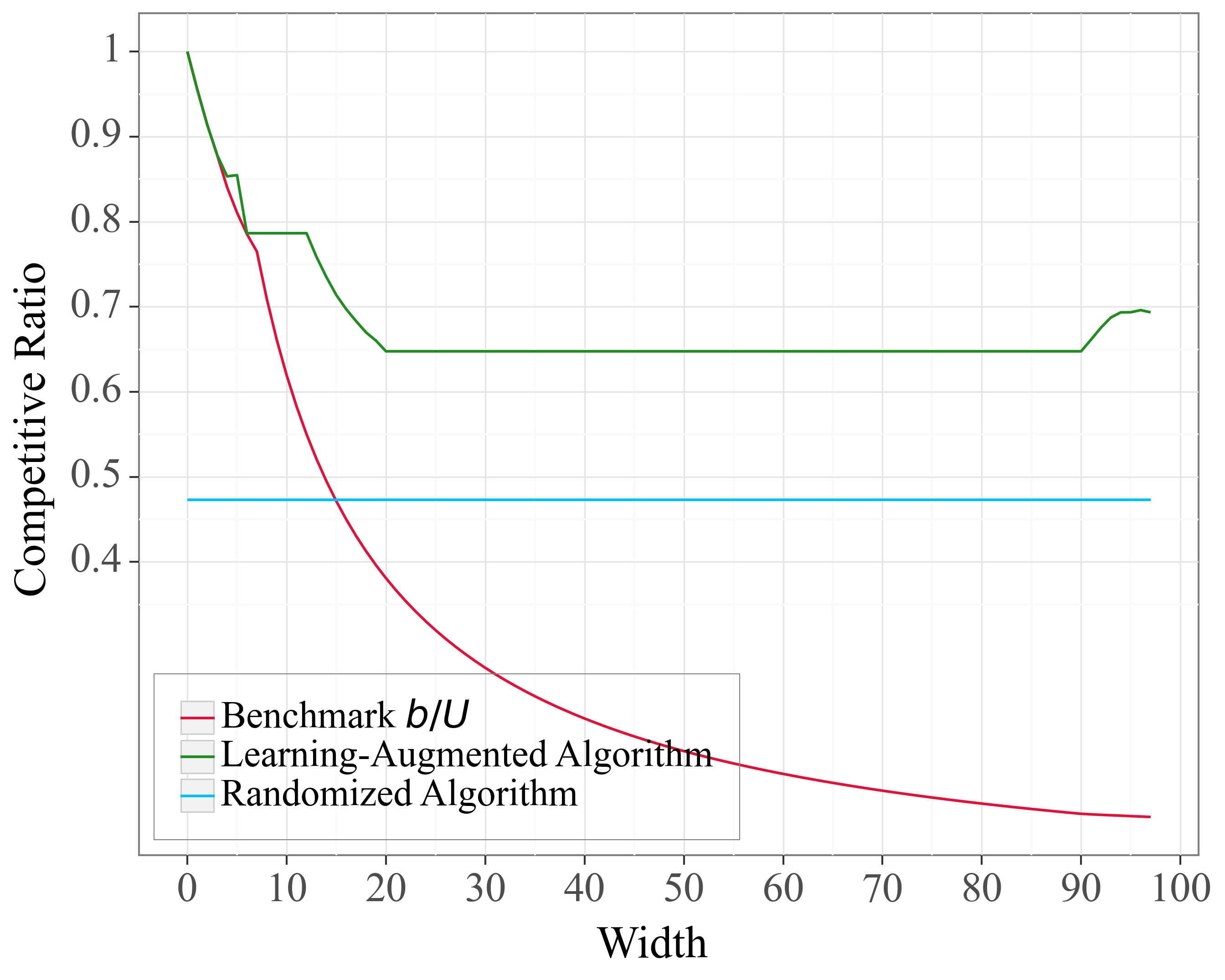}
        \vspace{-17pt}
        \caption{Scenario 3: $T=100$, $\tau^* = \text{Int}[0.1(T+b)]$}
        \label{fig:set22_c}
    \end{subfigure}
    \vspace{-8pt}
    \caption{Average competitive ratio under learning-augmented setting with $b=3$.}
    \label{fig:set22}
\end{figure*}

In this section, we examine the performances of the algorithms under the learning-augmented setting where $\tau^*$ is small. Specifically, we set the number of risk occurrences $\tau^* = \text{Int}[0.2(T+b)]$ for scenarios with horizon lengths $T = 22$ and $T=100$, and $\tau^* = \text{Int}[0.1(T+b)]$ for scenario $T = 100$. Figure \ref{fig:set22} presents the average competitive ratio against a range of prediction interval widths. 


\subsection{Budget Utilization by Each Algorithm}
\label{app:results_budget}
\begin{figure*}[ht!]
    \centering
    \begin{subfigure}[b]{0.31\linewidth}
        \centering
        \includegraphics[width=\linewidth,keepaspectratio]{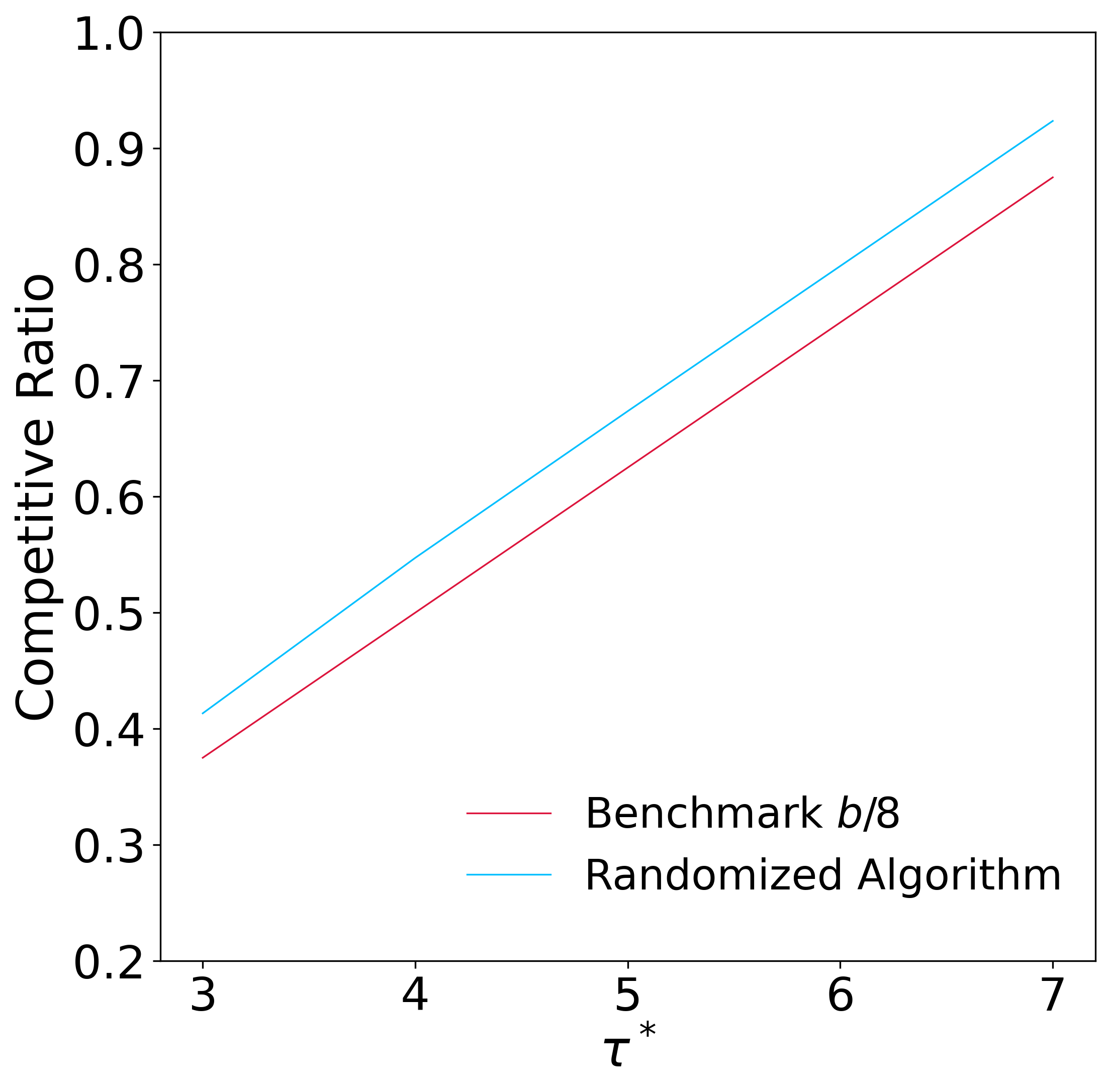}
        \vspace{-17pt}
        \caption{Scenario 1: $T=8$}
        \label{fig:set12_a}
    \end{subfigure}
    \hfill
    \begin{subfigure}[b]{0.31\linewidth}
        \centering
        \includegraphics[width=\linewidth,keepaspectratio]{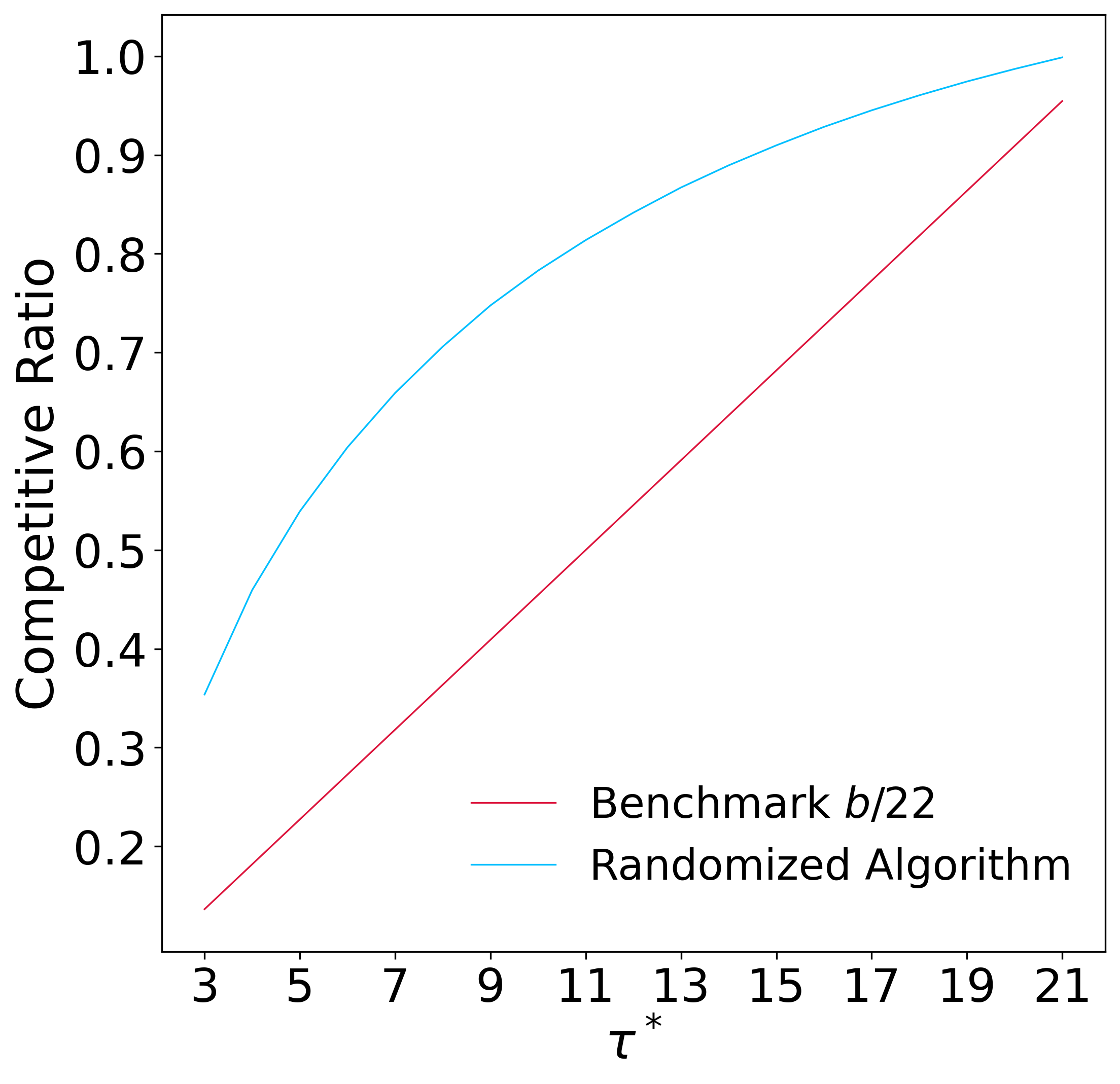}
        \vspace{-17pt}
        \caption{Scenario 2: $T=22$}
        \label{fig:set12_b}
    \end{subfigure}
    \hfill
    \begin{subfigure}[b]{0.31\linewidth}
        \centering
        \includegraphics[width=\linewidth,keepaspectratio]{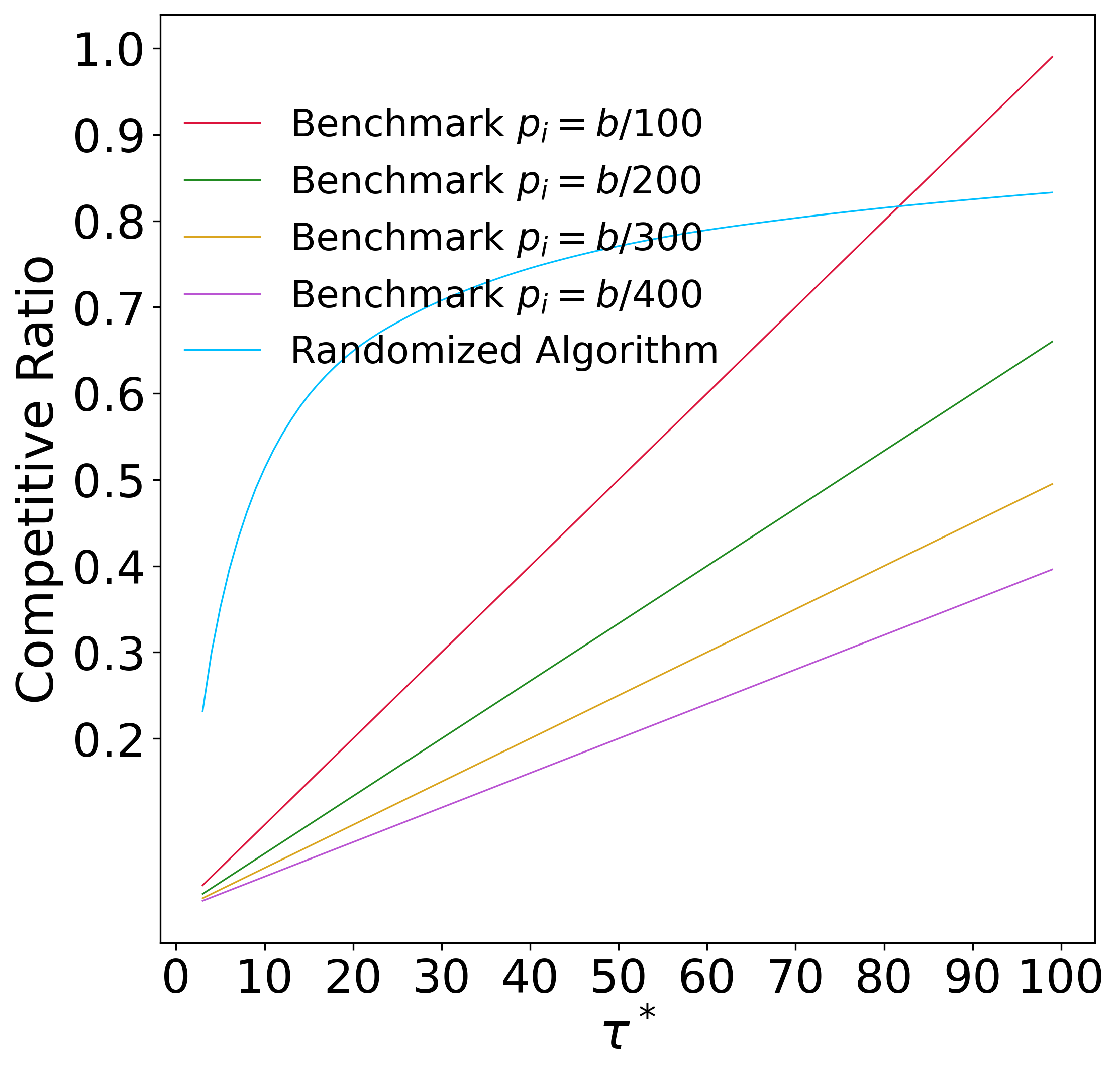}
        \vspace{-17pt}
        \caption{Scenario 3: $T=100$}
        \label{fig:set12_c}
    \end{subfigure}
    \vspace{-8pt}
    \caption{Average competitive ratio under non-learning augmented setting with $b=3$.}
    \label{fig:set12}
\end{figure*}

\begin{figure*}[ht!]
    \centering
    \begin{subfigure}[b]{0.31\linewidth}
        \centering
        \includegraphics[width=\linewidth,keepaspectratio]{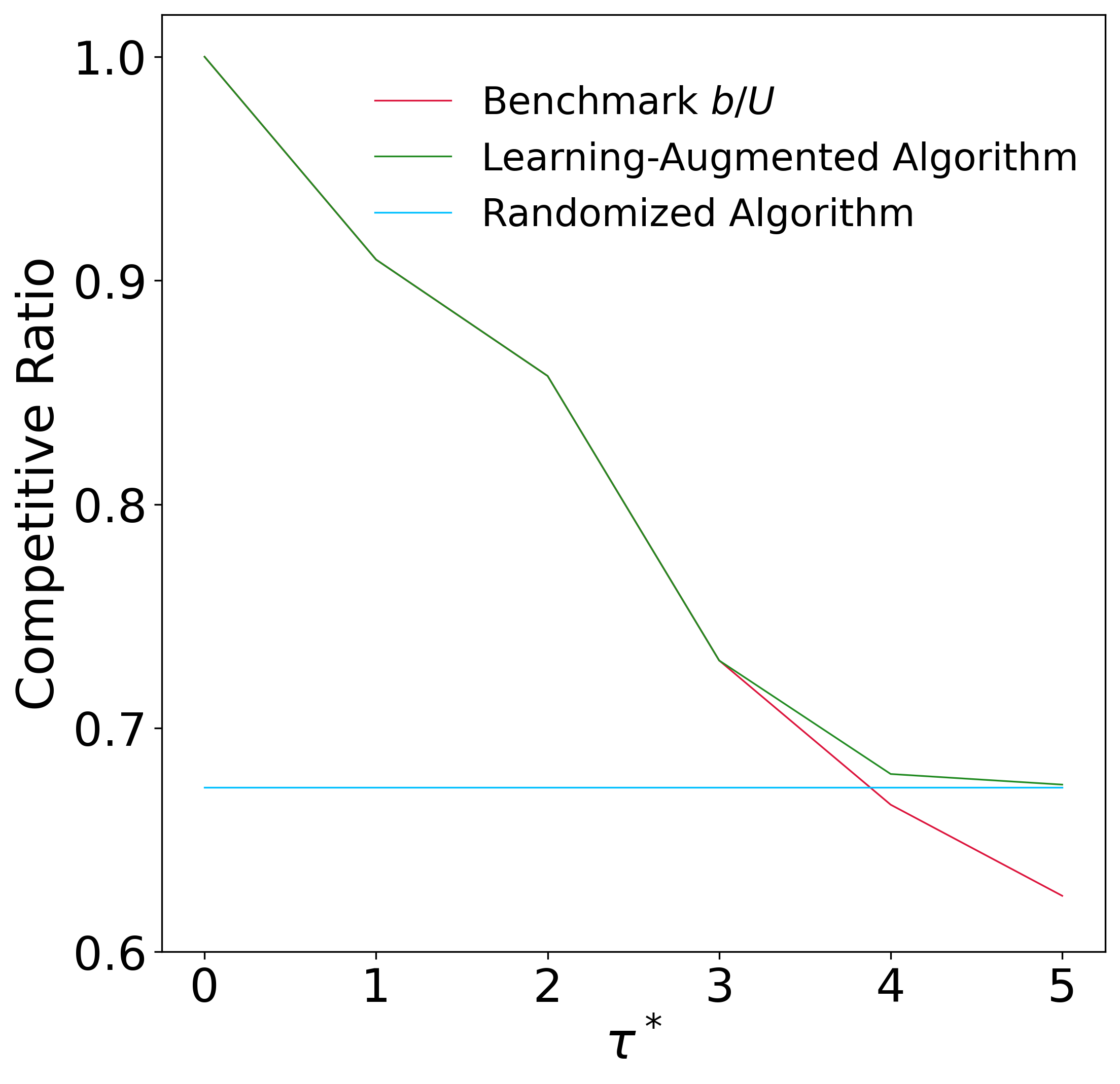}
        \vspace{-17pt}
        \caption{Scenario 1: $T=8$}
        \label{fig:set23_a}
    \end{subfigure}
    \hfill
    \begin{subfigure}[b]{0.31\linewidth}
        \centering
        \includegraphics[width=\linewidth,keepaspectratio]{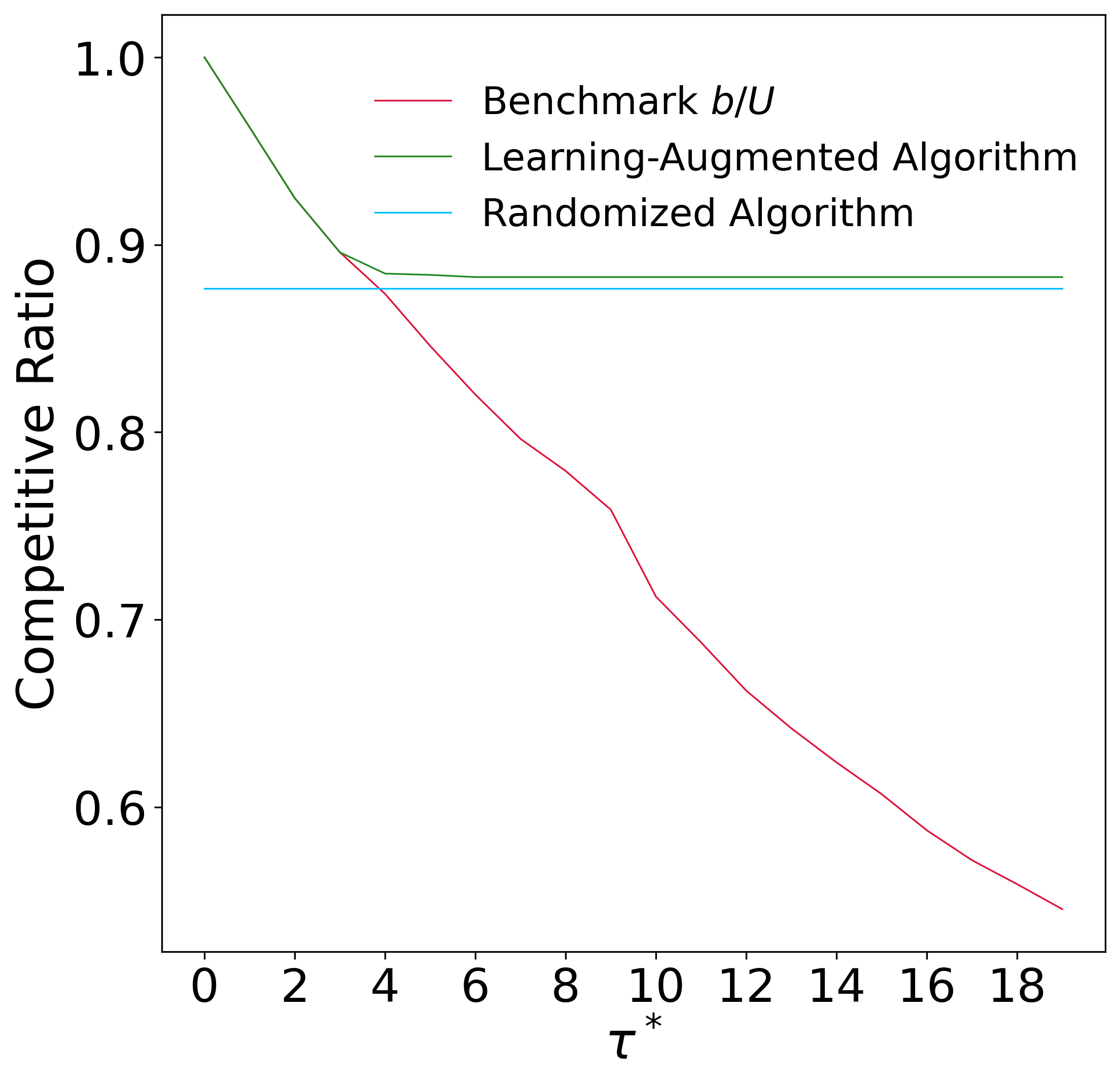}
        \vspace{-17pt}
        \caption{Scenario 2: $T=22$}
        \label{fig:set23_b}
    \end{subfigure}
    \hfill
    \begin{subfigure}[b]{0.31\linewidth}
        \centering
        \includegraphics[width=\linewidth,keepaspectratio]{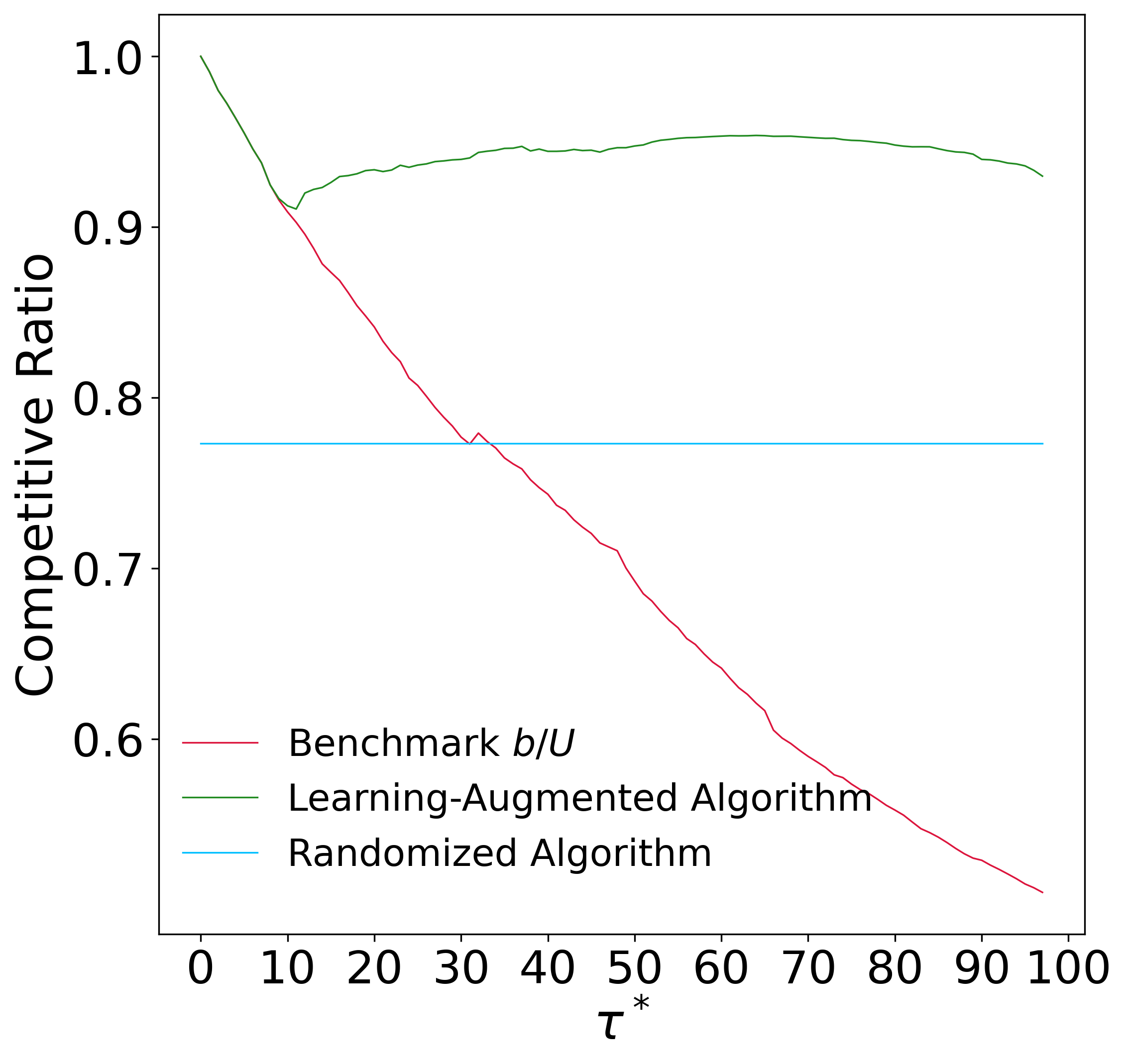}
        \vspace{-17pt}
        \caption{Scenario 3: $T=100$}
        \label{fig:set23_c}
    \end{subfigure}
    \caption{Average competitive ratio under learning-augmented setting with $b=3$.}
    \label{fig:set23}
\end{figure*}

To assess the budget utilization by each algorithm, we eliminate the penalty term from the objective in Problem~\ref{raw_opt}. Figures~\ref{fig:set12} and~\ref{fig:set23} display the average competitive ratios in scenarios without and with learning augmentation, respectively. We note that in Figure~\ref{fig:set12} (middle), when $\tau^*=22$, the competitive ratio slightly exceeds 1. This is attributed to our algorithm utilizing a slightly higher budget in expectation. We provide detailed insights into this observation in Section 1 of the Supplementary Material, where we demonstrate that the worst-case budget spent is about $1.047 b_k$, slightly surpassing the allocated budget.


\section{Additional Results on HeartSteps V1 Study}
\label{app:heartsteps}
Our research is inspired by the Heartsteps V1 mobile health study, which aims to enhance physical activity among sedentary individuals \citep{klasnja2019efficacy}. The study involved 37 participants over a follow-up period of six weeks,  gathering detailed data on step counts on a minute-by-minute basis.  To ensure the reliability of the step count data, our analysis was restricted to the hours from 9 am to 9 pm, with a decision time frequency set at five-minute intervals \citep{liao2018a}. This led to the accumulation of $1585$ instances of 12-hour user-days, with $T=144$ decision times per day.

At each decision time $t$,we define the risk variable $R_t$ with a binary classification: $R_t = 1$ indicates a sedentary state, identified by recording fewer than 150 steps in the prior 40 minutes, and $R_t=0$ signifies a non-sedentary state. Additionally, the availability for intervention, $I_t$, is contingent on recent messaging activity: if the user has received an anti-sedentary message within the preceding hour, $I_t$ is set to 0; otherwise, it is set to 1. We want to distribute $b=1.5$ interventions over available sedentary times each day. 

We implement four algorithms: our randomized and learning-augmented algorithms (Algorithms \ref{alg:rand} and \ref{alg:pred}, respectively), the SeqRTS strategy proposed by \citet{liao2018a}, and a benchmark method ($b/U$). 
Rather than devising a tailored prediction model, we generate prediction intervals by randomly selecting from a range of $[2,144]$, which contains $\tau^*$, with intervals of varying widths. This approach allows us to assess the performance of different algorithms under varying qualities of forecast accuracy.

We adopt the SeqRTS method to include prediction intervals, ensuring a balanced comparison with our algorithms. At the start of each user day, a number is randomly selected from the interval $[L,U]$ to estimate the number of available risk times. Should the budget be exhausted before allocating for all available risk times, a minimum probability of $1\times 10^{-6}$ is assigned to the remaining times. For additional information on the SeqRTS method, readers are referred to \citet{liao2018a}.

\begin{figure}[ht!]
    \centering
        \centering
        \includegraphics[width=0.4\linewidth,keepaspectratio]{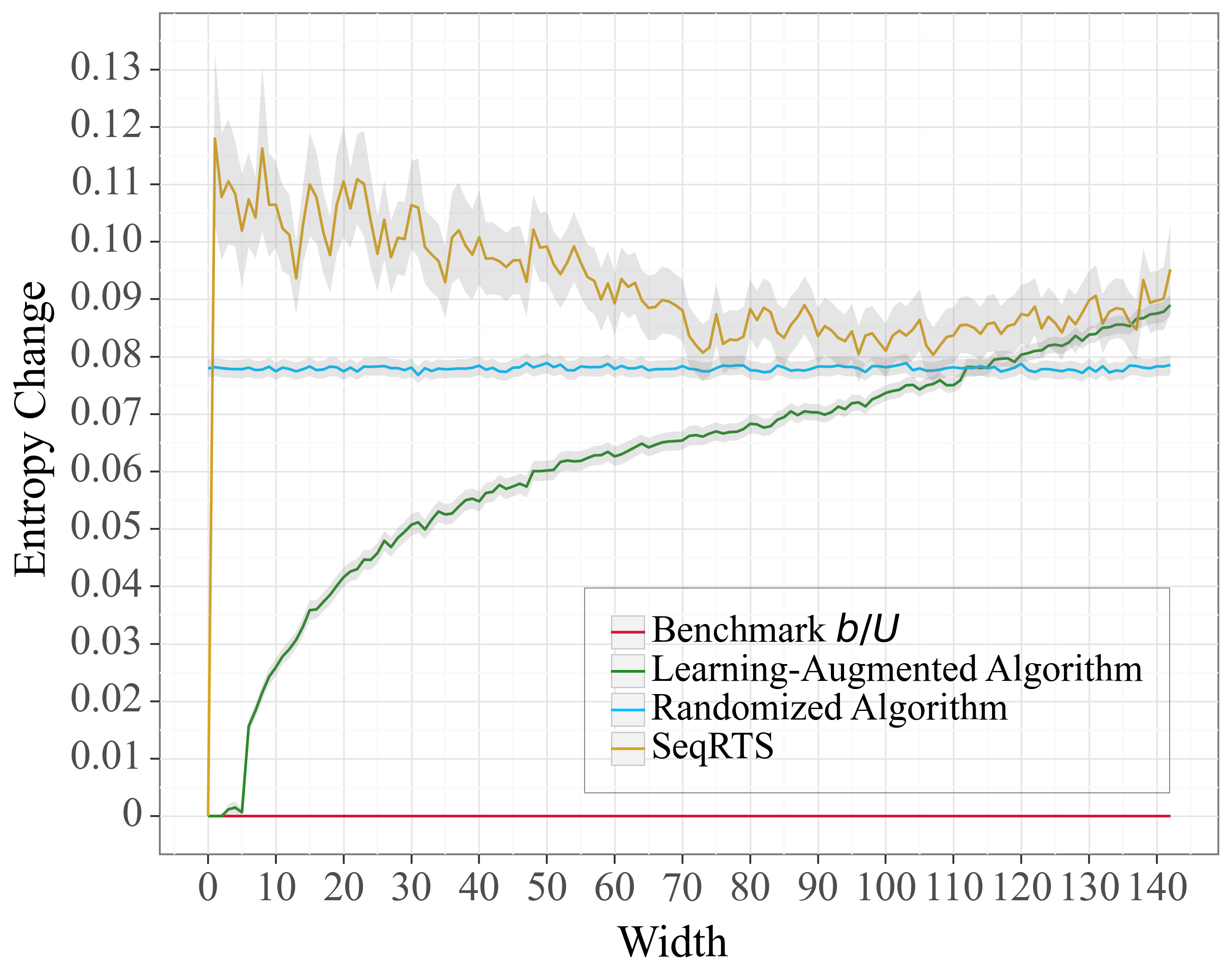}
        \vspace{-10pt}
    \caption{Average entropy change across user days under various prediction interval widths on HeartSteps V1 dataset. The shaded area indicates the $\pm 1.96$ standard error bounds across user days.}
            \label{fig:heart_b}
\end{figure}

Figure \ref{fig:heart_b} illustrates the average entropy change across user days. It is evident that SeqRTS exhibits the highest entropy change, suggesting non-uniform distribution behavior. In contrast, our learning-augmented algorithm demonstrates superior uniformity, outperforming the randomized algorithm. The benchmark method records an entropy of zero, attributed to its conservative strategy of assigning a constant probability of $b/U$.

\begin{figure}[ht!]
    \centering
    \begin{subfigure}[b]{0.4\linewidth}
        \centering
        \includegraphics[width=\linewidth,keepaspectratio]{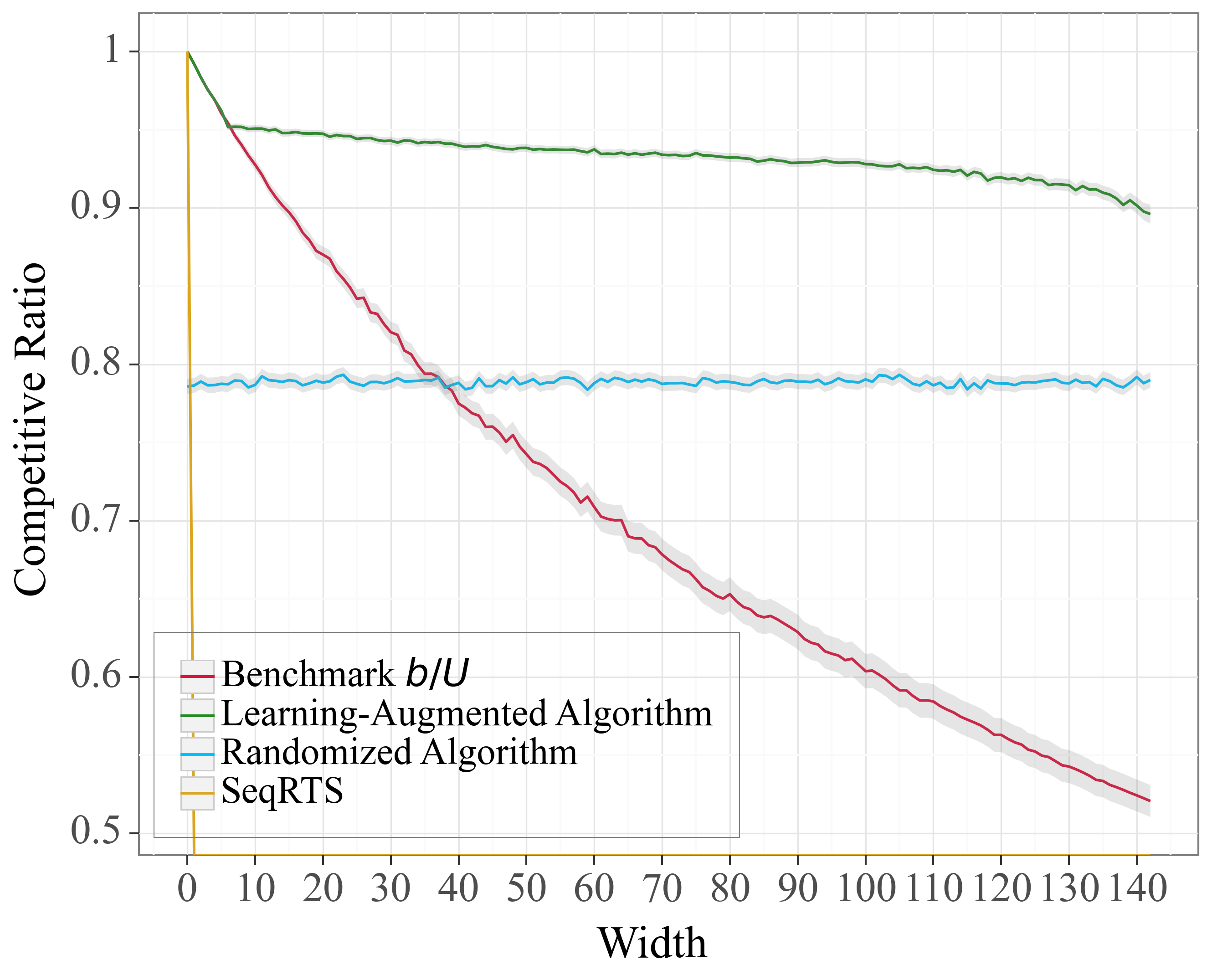}
        \vspace{-17pt}
        \caption{Competitive ratio}
        \label{fig:heart03_a}
    \end{subfigure}
    \begin{subfigure}[b]{0.4\linewidth}
        \centering
        \includegraphics[width=\linewidth,keepaspectratio]{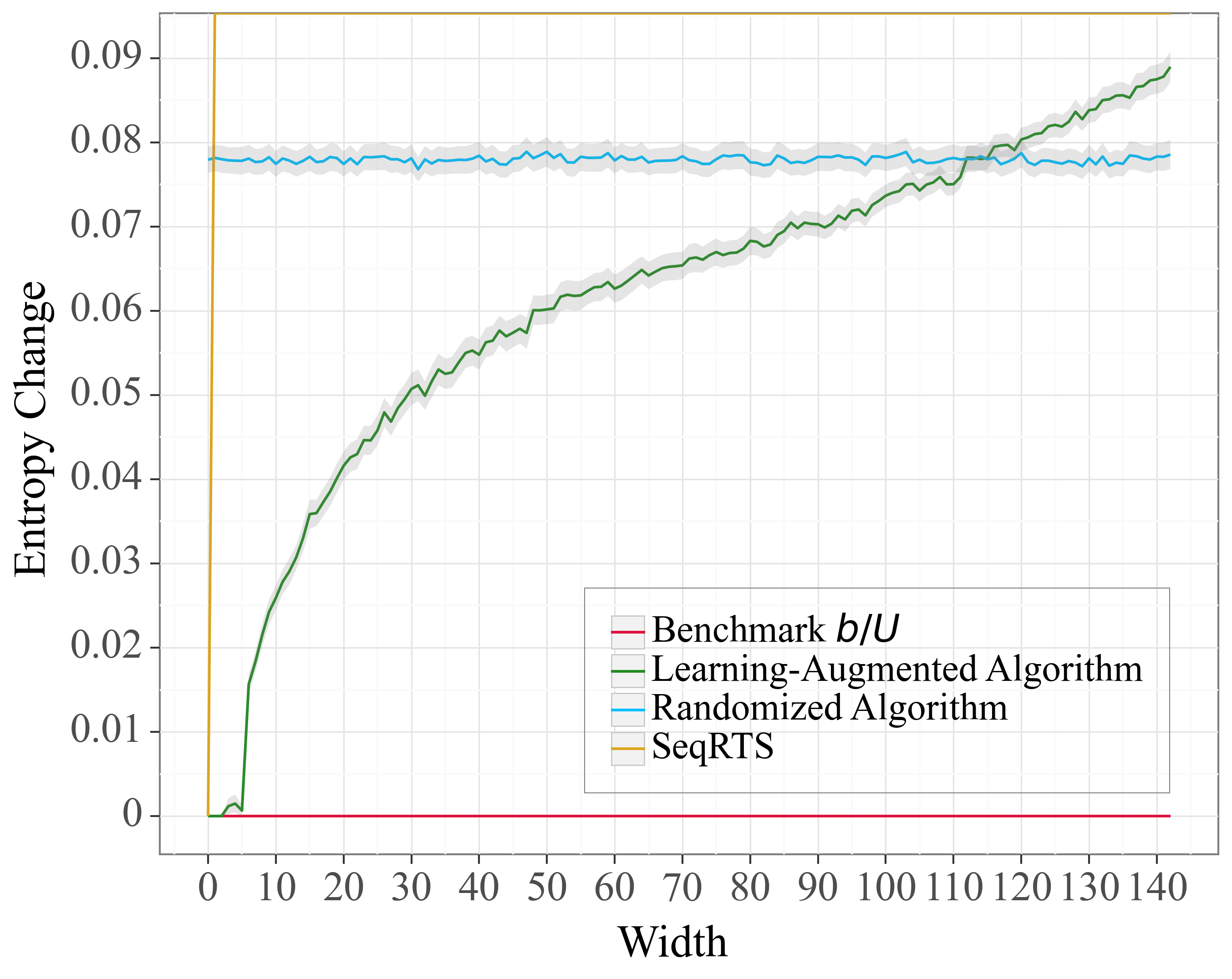}
        \vspace{-17pt}
        \caption{Entropy Change}
        \label{fig:heart03_b}
    \end{subfigure}
    \vspace{-8pt}
    \caption{Average competitive ratio and entropy change across user days under various prediction interval widths on the HeartSteps V1 dataset. The shaded area represents the $\pm 1.96$ standard error bounds across user days. \textit{Note}: For SeqRTS, a minimum probability of $0$ is assigned to the remaining times when the budget is exhausted.}
    \label{fig:heart03}
\end{figure}

Figure \ref{fig:heart03} shows the average competitive ratio and entropy change across user days, considering the scenario where SeqRTS assigns a minimum probability of $0$ to remaining risk times once the budget is depleted. Owing to the Penalization term \ref{eq:penalty}, this results in the objective function being negative infinity and the entropy change reaching infinity.

\end{document}